\documentclass{article}

\usepackage{arxiv}

\usepackage[utf8]{inputenc} % allow utf-8 input
\usepackage[T1]{fontenc}    % use 8-bit T1 fonts
\usepackage{hyperref}       % hyperlinks
\usepackage{url}            % simple URL typesetting
\usepackage{booktabs}       % professional-quality tables
\usepackage{amsfonts}       % blackboard math symbols
\usepackage{nicefrac}       % compact symbols for 1/2, etc.
\usepackage{microtype}      % microtypography
\usepackage{adjustbox}
\usepackage{textcomp}       % 支持特殊符号

\usepackage{lipsum}         % Can be removed after putting your text content
\usepackage{graphicx}
\usepackage[numbers,super]{natbib}
\usepackage{doi}
\usepackage{pifont}  

\usepackage{longtable}    % 长表格
\usepackage{booktabs}     % 表格线条
\usepackage{multirow}     % 合并行
\usepackage{array}        % 表格列类型
\usepackage{ragged2e}     % 文本对齐

\usepackage{algorithm}      % 算法环境
\usepackage{algpseudocode}  % 算法伪代码样式
\usepackage{amsmath}        % 数学符号
\usepackage{amssymb}        % 数学符号扩展
\usepackage{array}
\usepackage{makecell}

\usepackage{cleveref}       % smart cross-referencing

\title{A Study on Individual Spatiotemporal Activity Generation Method Using MCP-Enhanced Chain-of-Thought Large Language Models}

\date{}

\newif\ifuniqueAffiliation
\uniqueAffiliationtrue

\ifuniqueAffiliation % Standard variant of author block
\author{ Yu Zhang \\
	College of Architecture and Urban Planning\\
	Tongji University\\
	Shanghai 200092, China \\
	\texttt{2330332@tongji.edu.cn} \\
	\And
	Yang Hu \\
	College of Architecture and Urban Planning\\
	Tongji University\\
	Shanghai 200092, China \\
	\texttt{huyang@tongji.edu.cn} \\
	\And
	De Wang \\
	College of Architecture and Urban Planning, Tongji University\\
	Digital Planning Technology Research Center, Shanghai Tongji Urban Planning \& Design Institute Co., Ltd.\\
	Shanghai 200092, China \\
	\texttt{dewang@tongji.edu.cn} \\
}
\else
\usepackage{authblk}

\author[1]{Yu Zhang\thanks{\texttt{zhangyu@tongji.edu.cn}}}
\author[1]{Yang Hu\thanks{\texttt{huyang@tongji.edu.cn}}}
\author[1,2]{De Wang\thanks{\texttt{dewang@tongji.edu.cn}}}

\affil[1]{College of Architecture and Urban Planning, Tongji University, Shanghai 200092, China}
\affil[2]{Digital Planning Technology Research Center, Shanghai Tongji Urban Planning \& Design Institute Co., Ltd., Shanghai 200092, China}
\fi

\renewcommand{\shorttitle}{\textit{MCP-Enhanced CoT for Individual Spatiotemporal Behavioral Generation}}

\hypersetup{
   pdftitle={A Study on Individual Spatiotemporal Behavior Generation Method Using MCP-Enhanced Chain-of-Thought Large Language Models},
   pdfsubject={cs.AI, cs.HC, cs.CY},
   pdfauthor={Yu Zhang, Yang Hu, De Wang},
   pdfkeywords={Large Language Models, Chain-of-Thought Reasoning, MCP Protocol, Spatiotemporal Behavior Simulation, Urban Computing, Activity Trajectory Generation, Individual Behavior Modeling, Spatial Data Mining},
}

\begin{document}
\maketitle

\begin{abstract}
	\quad Human spatiotemporal behavior simulation is important for research in urban planning and related fields, yet traditional rule-based and statistical approaches face significant limitations including high computational costs, limited generalizability, and poor scalability. While large language models (LLMs) show promise as "world simulators," they face important challenges in spatiotemporal reasoning including limited spatial cognition, lack of physical constraint understanding, and group homogenization tendencies. This paper introduces a framework that integrates chain-of-thought (CoT) reasoning with Model Context Protocol (MCP) to enhance LLMs' capability in simulating and generating spatiotemporal behaviors that correspond with validation data patterns. The methodology combines human-like progressive reasoning through a five-stage cognitive framework with comprehensive data processing capabilities via six specialized MCP tool categories: temporal management, spatial navigation, environmental perception, personal memory, social collaboration, and experience evaluation. Experiments in Shanghai's Lujiazui district validate the framework's effectiveness across 1,000 generated samples. Results indicate correspondence with real mobile signaling data, achieving generation quality scores ranging from 7.86 to 8.36 across different base models in controlled experimental conditions. Parallel processing experiments demonstrate efficiency improvements under tested configurations, with generation times decreasing from 1.30 to 0.17 minutes per sample when scaling from 2 to 12 processes on dedicated hardware, thereby demonstrating the framework's potential for large-scale deployment.
	
	\quad This work contributes to the integration of CoT reasoning with MCP for urban behavior modeling, advancing LLMs applications in urban computing and providing a practical approach for synthetic mobility data generation in data-constrained environments. The framework offers a basis for exploring potential applications in smart city planning, transportation forecasting, participatory urban design, and related domains.\footnote{\textsuperscript{*}For detailed code and additional information, please visit the code repository: \url{https://github.com/ZYY799/spatiotemporal-activity-generation-mcp-cot}. The trial system web platform is currently in the development and debugging phase.}\textsuperscript{*}
\end{abstract}

% keywords can be removed

\section{Introduction}
Human spatiotemporal behavior\footnote{\textsuperscript{*}In this study, "spatiotemporal behavior" refers to human activity-travel patterns characterized by: (1) \textit{spatiotemporal coordinates}: locations, timing, duration, and spatial-temporal sequences of activities and movements; and (2) \textit{behavioral dimensions}: individual characteristics (age, household composition, socioeconomic status), activity purposes and types, travel companions, route trajectories, transportation mode choices, subjective perceptions and preferences, post-activity evaluations, and future planning strategies.}\textsuperscript{*} constitutes the foundational data for urban research,  proving important for multiple domains including urban planning, transportation management, commercial site selection, and epidemic control. Accurately understanding and predicting individual spatiotemporal behavior patterns in cities enables optimization of urban functional layouts, improvement of resource utilization efficiency, and enhancement of public service quality~\cite{wang2020review,haraguchi2022human}. However, high-quality spatiotemporal behavioral data acquisition remains challenging~\cite{hamdi2022spatiotemporal} as conventional survey methods are constrained by cost and sample limitations, whereas LBS-based and mobile signaling approaches present privacy vulnerabilities and insufficient semantic granularity. Although rule-based behavioral simulation and statistical model-driven spatiotemporal behavior generation methods have partially mitigated data scarcity issues~\cite{pappalardo2018data}, these approaches generally rely on substantial prior knowledge and produce overly simplified behavioral patterns that fail to capture the complexity and diversity inherent of human activities~\cite{fuchs2023modeling}.

In recent years, Large Language Models (LLMs) have demonstrated significant advances in knowledge integration and logical reasoning. Through pre-training on massive text corpora, these models have accumulated extensive world knowledge, including understanding of human behavioral patterns and social activities. Consequently, LLMs possess the potential to serve as "world simulators," capable of simulating complex real-world systems and human behaviors. However, LLMs face several important challenges when applied to urban spatiotemporal behavior simulation: First, limited spatial cognitive abilities make it difficult to accurately understand and express geographic coordinates and spatial relationships~\cite{yamada2023evaluating}. Second, insufficient understanding of real-world physical constraints may lead to generation of behavior sequences that violate spatiotemporal continuity~\cite{li2024geo}. Third, closed training environments restrict the model's ability to acquire and update external dynamic information. Fourth, models exhibit pronounced group homogenization tendencies, with generated content often reflecting "average" and standard behavioral patterns while struggling to capture population heterogeneity and individual unique behavioral patterns~\cite{moon2024homogenizing}.Fifth, LLMs generate different outputs each time due to random sampling, preventing reproducible results and creating challenges for interpretability of generated behavioral patterns~\cite{doshi2017towards}. Additionally, generating behavior data for entire city populations presents significant computational resource challenges for large-scale applications.

To address these challenges, this paper develops an individual spatiotemporal behavior generation method that integrates MCP-enhanced chain-of-thought reasoning with LLMs. This method integrates two technical paradigms: Chain-of-Thought Reasoning and Model-Context Protocol (MCP), enabling LLMs to progressively reason through activity decision-making processes through structured reasoning while acquiring and processing precise spatiotemporal information through a unified protocol framework. MCP, as a more flexible and powerful context interaction mechanism compared to traditional function calling, can effectively integrate multi-source heterogeneous data, support complex conditional reasoning, and facilitate dynamic interaction between individual memory and group knowledge, thereby providing a novel technical approach for large-scale individual spatiotemporal behavior generation. Furthermore, the Lujiazui area of Shanghai is selected as an empirical case study to evaluate the effectiveness of the proposed method through comparative experiments.

The key innovations and contributions of this work include: (1) Proposing a novel spatiotemporal behavior generation method combining chain-of-thought reasoning and MCP, extending the application boundaries of large models in urban spatial computing; (2) Developing a comprehensive MCP framework with specialized tool collections for urban activity simulation, addressing limitations of LLMs in spatial precision and logical coherence; (3) Constructing large-scale, high-quality spatiotemporal behavioral knowledge bases and local knowledge repositories, establishing a systematic data training and validation framework; (4) Developing a parallel processing pipeline suitable for large-scale spatiotemporal behavior generation, addressing efficiency and consistency issues in batch generation; (5) Demonstrating through experimental validation the advantages of the proposed method in generating high-quality, realistic individual spatiotemporal behaviors, thus establishing a novel technical framework for human behavior simulation under data-limited conditions.

\section{Related Work}
\label{sec:Related Work}

\subsection{Agent-Based Human Behavior Simulation}
\label{sec:agent_based_simulation}
Agent-Based Modeling (ABM) has emerged as a valuable methodology for simulating individual behaviors and social systems. Within urban studies, ABM utilizes autonomous agents equipped with decision-making capabilities and defined interaction rules to generate macro-level phenomena from micro-level behaviors, thereby establishing a "bottom-up" framework for analyzing complex systems. Batty et al.~\cite{batty2001agent} established the foundation for applying ABM to urban pedestrian behavior simulation, demonstrating how simple behavioral rules could effectively reproduce collective movement patterns in complex urban environments. Building upon this foundation, traffic simulation platforms such as MATSim~\cite{w2016multi} have advanced this approach by integrating synthetic populations with transportation networks to achieve large-scale dynamic simulations of urban traffic flows, providing valuable empirical evidence for urban planning and traffic management.

Traditional ABM approaches, however, face two fundamental limitations. First, behavioral rule design relies extensively on expert knowledge, which constrains the model's ability to fully capture the inherent complexity, variability, and adaptability of human decision-making. Second, computational demands scale exponentially with simulation size, restricting practical applications to relatively small urban systems. In response to these challenges, scholars have increasingly incorporated machine learning techniques into ABM frameworks. Zheng et al.~\cite{zheng2018exploiting} developed an agent behavior model leveraging deep reinforcement learning, enabling agents to autonomously acquire decision strategies from empirical data rather than relying on predefined rules, thereby significantly enhancing adaptability to complex environments. Similarly, Liu et al.~\cite{liu2020multi} employed graph neural networks to model agent interactions, effectively capturing both the influence of social network structures on individual behaviors and the intricate mechanisms underlying inter-agent dynamics. Despite these advances, such approaches remain computationally intensive and data-hungry, requiring extensive labeled datasets for training while exhibiting limited generalization in data-sparse environments or novel scenarios.

Recent developments in large language model (LLM)-based agent simulation frameworks have unveiled notable possibilities. Park et al.~\cite{park2023generative} introduced the concept of Generative Agents, which harness LLMs to construct sophisticated memory, reflection, and decision-making mechanisms, enabling the simulation of highly realistic and diverse human behaviors. Complementing this work, frameworks such as AutoGen~\cite{wu2023autogen} and LangChain~\cite{topsakal2023creating} have established architectures for multi-agent collaboration, facilitating autonomous interaction and cooperation among LLM-driven agents. While these studies demonstrate the considerable promise of large language models for human behavior simulation, significant challenges persist in addressing the distinctive characteristics of urban spatiotemporal behaviors, including environmental constraints, multi-scale interactions, and long-term evolutionary dynamics.

\subsection{Activity Chain and Spatiotemporal Behavior Research}
\label{sec:activity_chain_research}
Activity chain simulation represents an important approach for understanding and predicting human daily behavior. Hägerstrand's~\cite{hagerstrand1970people} time geography framework emphasizes the impact of spatiotemporal constraints on human activities, providing a theoretical foundation that has inspired numerous empirical studies and model developments. Granovetter~\cite{granovetter1975human} analyzed the motivations and processes underlying activity selection from a sociological perspective, highlighting the roles of personal preferences, social roles, and lifestyles in activity decision-making.

With advances in location-aware technologies, data-driven spatiotemporal behavior research has made considerable progress. Chen et al.~\cite{chen2016promises} reviewed applications of big data and small data in human mobility analysis, demonstrating that mobile phone location data can be used to identify common activity patterns of urban residents. Widhalm et al.~\cite{widhalm2015discovering} effectively extracted stable activity clusters from multi-city mobile phone data by integrating activity timing, duration, and land-use characteristics; similarly, Ahas et al.~\cite{ahas2010using} analyzed the spatiotemporal distribution characteristics of different social groups using mobile phone positioning data, while Yin et al.~\cite{yin2021mining} discovered prevalent activity patterns of urban residents through mobile phone trajectory data. Nevertheless, these observation-based methods generally require extensive location data and face constraints related to data sampling frequency, accuracy, and coverage.

To overcome the limitations of observational data, various activity chain generation models have been proposed. Among these, CEMDAP~\cite{bhat2004comprehensive}, ALBATROSS~\cite{arentze2000albatross}, and TASHA~\cite{roorda2008validation} generate activity chains based on demographic characteristics by integrating rule-based systems with statistical models. Drchal et al.~\cite{drchal2019data} introduced DDAS, a deep generative model-based approach for activity chain generation that learns from extensive real-world activity data to produce activity sequences conforming to statistical distributions. Nevertheless, these models typically employ multi-stage decision processes: first determining activity types and temporal arrangements, then assigning activity locations, and finally planning transportation modes. Despite these advances, such methods tend to rely substantially on extensive labeled data, and the generated activity patterns often exhibit limited individual-specific characteristics, potentially constraining their ability to capture the full diversity and uniqueness of human behavior. Moreover, this staged linear decision framework may inadequately represent the dynamic trade-offs and interdependencies among various factors in actual human decision-making processes.

\subsection{Reasoning Capabilities and Applications of Large Language Models}
\label{sec:llm_reasoning}
Large Language Models (LLMs) have demonstrated impressive performance across numerous tasks through the pre-training and fine-tuning paradigm. Models such as ChatGPT and DeepSeek not only generate fluent natural language but also exhibit notable reasoning capabilities and extensive world knowledge. These capabilities emerge from extensive textual data processed during pre-training. Research indicates that LLMs appear to implicitly learn complex concepts, facts, and reasoning rules from such data~\cite{brown2020language}~\cite{xie2021explanation}.

Building upon these capabilities, scholars have begun exploring how to leverage LLMs' core training mechanism—next-token prediction—to directly model human activity sequences. This approach treats human mobility trajectories and activity chains as specialized sequential data, capturing inherent spatiotemporal correlations and mobility patterns through fine-tuned language models~\cite{liang2024exploring}. The Geo-Llama framework~\cite{li2024geo} established this approach by employing a visit-wise arrangement strategy that enables models to capture authentic mobility patterns without temporal ordering constraints. NextlocLLM~\cite{liu2024nextlocllm} further improved spatial relationship modeling by encoding continuous spatial coordinates rather than discrete location IDs. AgentMove decomposed mobility prediction into multiple subtasks, incorporating spatiotemporal memory modules to extract individual mobility patterns~\cite{feng2024agentmove}. These token prediction-based approaches demonstrate how LLMs can effectively model complex spatiotemporal behaviors, though their reasoning processes often exhibit limited transparency and interpretability.

In response to the interpretability challenges inherent in these spatiotemporal modeling approaches, Chain-of-Thought (CoT) reasoning emerges as a promising technique that helps mitigate this limitation by enhancing the reasoning transparency of LLMs. Wei et al.~\cite{wei2022chain} showed that guiding models to generate intermediate reasoning steps can significantly improve performance on mathematical problems and commonsense reasoning tasks. Subsequent research revealed that CoT reasoning benefits models of various scales, particularly during knowledge distillation, while Kojima et al.\cite{kojima2022large} further demonstrated that simple prompts such as "let's think step by step" can elicit reasoning capabilities, achieving zero-shot chain-of-thought reasoning. These explicit reasoning mechanisms provide valuable foundations for developing more interpretable and controllable spatiotemporal decision-making systems, helping to resolve the transparency limitations observed in current LLM-based mobility modeling frameworks.

Complementing these interpretability advances, the Model-Context Protocol (MCP) represents a structured interaction paradigm that enhances LLMs' understanding of external environments. In contrast to the Toolformer model proposed by Schick et al.~\cite{schick2023toolformer}, which primarily focuses on discrete tool invocation, MCP emphasizes the establishment of unified contextual interaction protocols. These protocols facilitate integration between LLMs and diverse external systems, knowledge repositories, and computational environments. Through the embedding of structured instructions and responses within contextual frameworks, MCP enables LLMs to better comprehend and process complex external environments, thereby providing a suitable foundation for handling spatiotemporal information and decision logic.

These foundational capabilities and methodological advances have enabled diverse applications of LLMs in urban computing across multiple research domains. Zhang et al.\cite{zhang2024trafficgpt} introduced the TrafficGPT framework, which integrates LLMs with domain-specific traffic management expertise to support observation, analysis, and processing of traffic data while facilitating decision support. Similarly, Zhou et al.\cite{zhou2024large} developed a multi-agent collaborative framework based on LLMs for participatory urban planning, generating land-use plans that consider diverse stakeholder needs through simulated interactions between urban planners and residents. Furthermore, Ning and Liu~\cite{ning2024urbankgent} presented UrbanKGent, a unified framework that combines LLMs with urban knowledge graphs for knowledge graph construction. This methodology extracts relevant knowledge from multi-source urban data through heterogeneous perception, geospatial-injected instruction generation, and tool-enhanced iterative trajectory optimization, thereby supporting urban problem diagnosis and solution recommendation.

Despite these promising developments, current applications primarily emphasize textual understanding and generation capabilities, with limited exploration of precise spatiotemporal data processing, particularly at the microscopic individual behavioral level in urban research contexts. This limitation highlights the need for developing targeted methodological approaches that better integrate LLMs' reasoning capabilities with spatiotemporal modeling requirements.

\subsection{Large-scale Parallel Processing Frameworks}
Large-scale spatiotemporal behavior generation requires efficient parallel processing frameworks to handle the computational demands inherent in such applications. While traditional parallel computing frameworks such as MapReduce~\cite{dean2008mapreduce} and Spark~\cite{zaharia2010spark} provide foundational big data processing capabilities, they encounter notable limitations when applied to LLM-based applications that require complex reasoning and interactive processing workflows. Ray~\cite{moritz2018ray} offers a distributed computing framework specifically designed to support parallel inference and batch processing for LLMs. Advanced optimization tools such as DeepSpeed~\cite{rasley2020deepspeed} and vLLM~\cite{kwon2023efficient} enhance LLM inference efficiency through sophisticated techniques including continuous batching, tensor parallelism, and PagedAttention mechanisms, achieving notable throughput improvements for large-scale deployment scenarios.

At the system coordination level, multi-agent collaboration frameworks including LangGraph and AgentForce have been developed to provide distributed architectures for coordinating multiple agents in LLM-driven systems. AgentForce, which is built upon the Atlas reasoning engine, enables autonomous reasoning and decision-making capabilities while effectively managing complex workflows in large-scale agent applications. These technological advances provide valuable insights and architectural references for developing the large-scale spatiotemporal behavior generation approach proposed in this research.

\begin{figure}[htbp]
    \centering
    \includegraphics[width=1\textwidth]{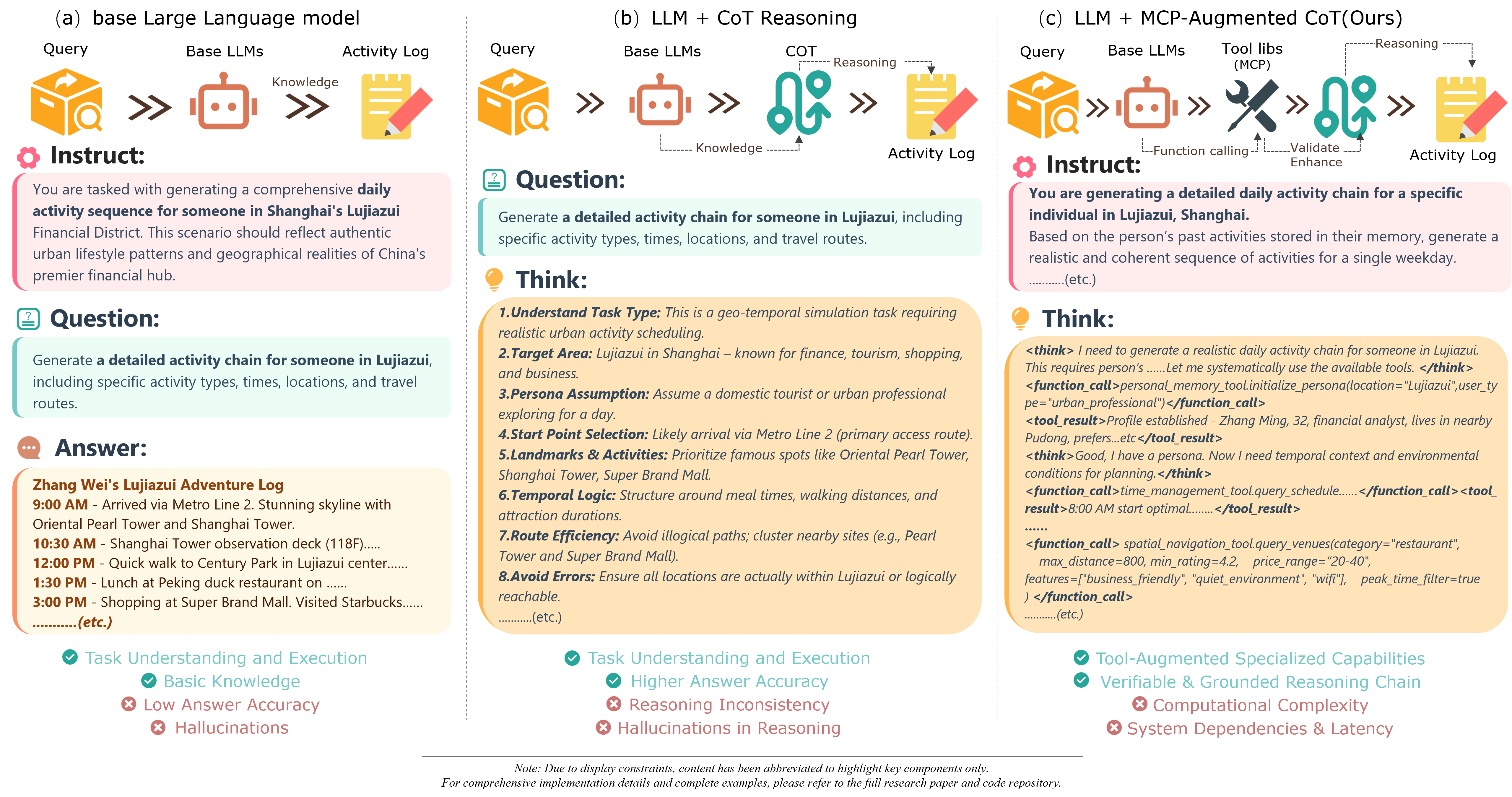}
    \caption{Evolution and Comparison of LLM-based Urban Activity Generation Methods}
    \label{fig:EvolutionandComparisonofLLM-based}
\end{figure}

\textbf{In summary}, while traditional ABM methods, activity chain simulation techniques, and LLMs have each achieved notable progress in their respective domains, research that effectively integrates these approaches for large-scale urban individual spatiotemporal behavior generation remains relatively limited. Important challenges persist, including: how to enhance the spatiotemporal reasoning capabilities of LLMs, how to construct key application components (tools) for spatiotemporal behavior through MCP protocols, how to design efficient parallel processing frameworks supporting batch generation, and how to ensure the balance between individual diversity and collective consistency in generated spatiotemporal behavior. This research aims to bridge this gap by combining CoT reasoning with MCP-based tool construction to build a system capable of efficiently generating realistic and reasonable urban individual spatiotemporal behavior(Figure \ref{fig:EvolutionandComparisonofLLM-based}).

\section{Theoretical Foundation and Methodology Design}
\label{sec:Theoretical Foundation and Methodology Design}

\subsection{Theoretical Framework}

\begin{figure}[htbp]
    \centering
    \includegraphics[width=0.7\textwidth]{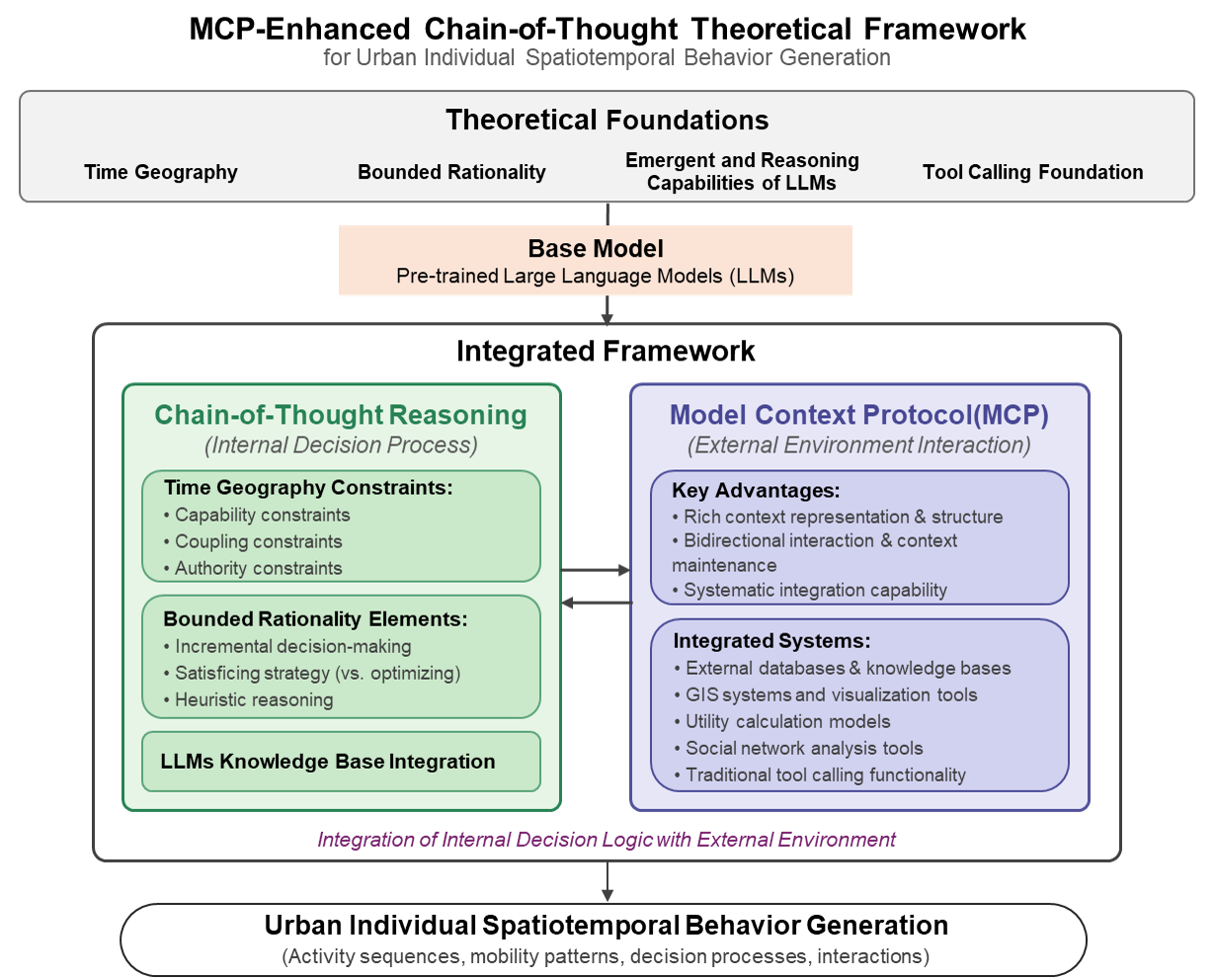}
    \caption{MCP-Enhanced Chain-of-Thought Theoretical Framework}
    \label{fig:MCP-EnhancedChain-of-ThoughtTheoreticalFramework}
\end{figure}

A theoretical framework for urban individual spatiotemporal behavior generation incorporates MCP-enhanced CoT reasoning (Figure \ref{fig:MCP-EnhancedChain-of-ThoughtTheoreticalFramework}). This approach synthesizes time geography, bounded rationality decision theory, the emergent capabilities of LLMs, and MCP, creating a comprehensive individual spatiotemporal behavior simulation and generation system.

Time geography provides the fundamental spatiotemporal constraint theory for this approach. According to Hägerstrand's classical theory~\cite{hagerstrand1970people}, individual activities are subject to three fundamental constraints: capability constraints, coupling constraints, and authority constraints. These constraints together form an individual's spatiotemporal accessibility range, known as the "spatiotemporal prism." Within this model, these constraints are encoded as contextual environment parameters through MCP, guiding LLMs to generate activity sequences that conform to realistic physical limitations.

Bounded rationality decision theory provides the cognitive modeling foundation for the approach. Unlike the perfect rationality assumption in traditional economics, Simon's bounded rationality theory~\cite{simon1955behavioral} posits that decision-makers in real-world environments are constrained by cognitive capacity, information acquisition, and processing time, often adopting "satisficing" rather than "optimizing" strategies. This model incorporates the core tenets of bounded rationality—incremental decision-making, satisficing solution search, and heuristic reasoning—into the CoT design, thereby simulating human decision-making logic in complex spatiotemporal environments, including spatial decisions, activity selection, travel mode choice, and route planning.

The emergent capabilities of LLMs provide the technical and knowledge foundation for the entire system. These pre-trained models trained on large-scale text corpora have developed implicit representations of world knowledge, including human behavioral patterns, social activity regularities, and geographical spatial understanding. CoT reasoning activates the model's step-by-step reasoning capabilities, enabling systematic consideration of multiple factors and constraints while making the cognitive process transparent and revealing implicit knowledge explicitly.

MCP serves as the key innovation of this approach, providing enhanced interaction capabilities beyond traditional tool calling. Through MCP, the system enables the construction of structured spatiotemporal environment representations and supports multi-round reasoning processes, facilitating real-time external environment queries, spatial information retrieval, external database connections, and computational tool integration to create authentic contextual environments for individual spatiotemporal behavior simulation.

Compared to existing research, this approach establishes an integrated simulation mechanism spanning from individual internal decision-making to external environmental interaction by integrating two core components: CoT reasoning and MCP. CoT reasoning simulates the internal decision-making processes of individuals, while MCP facilitates the interaction between individuals and external environments. These components work synergistically to overcome the limitations of traditional methods in terms of authenticity and generalizability.

\subsection{Methodology Design}

\begin{figure}[htbp]
    \centering
    \begin{minipage}{0.42\textwidth}
        \centering
        \includegraphics[width=\linewidth]{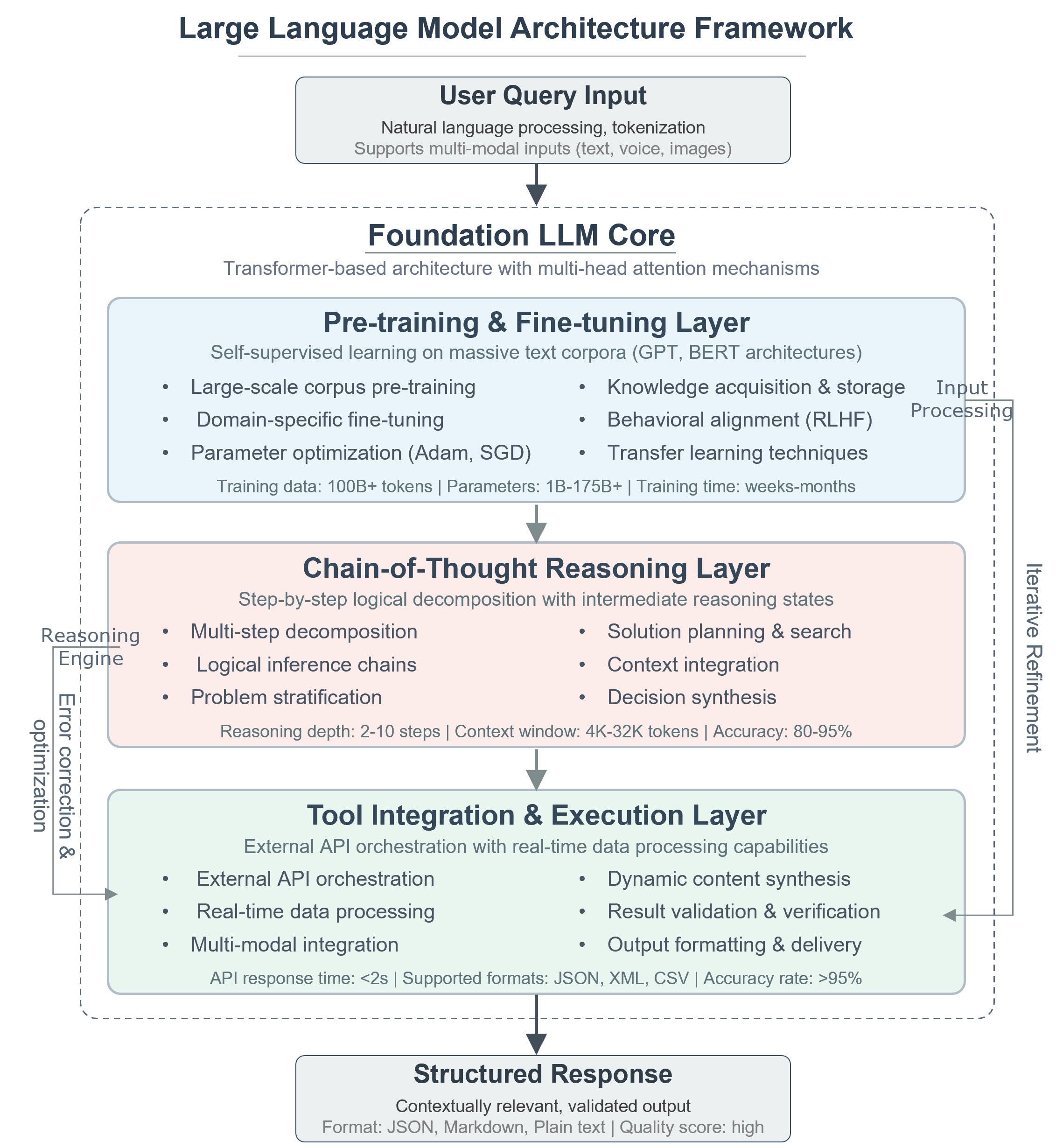}
    \end{minipage}%
    \hfill
    \begin{minipage}{0.55\textwidth}
        \centering
        \includegraphics[width=\linewidth]{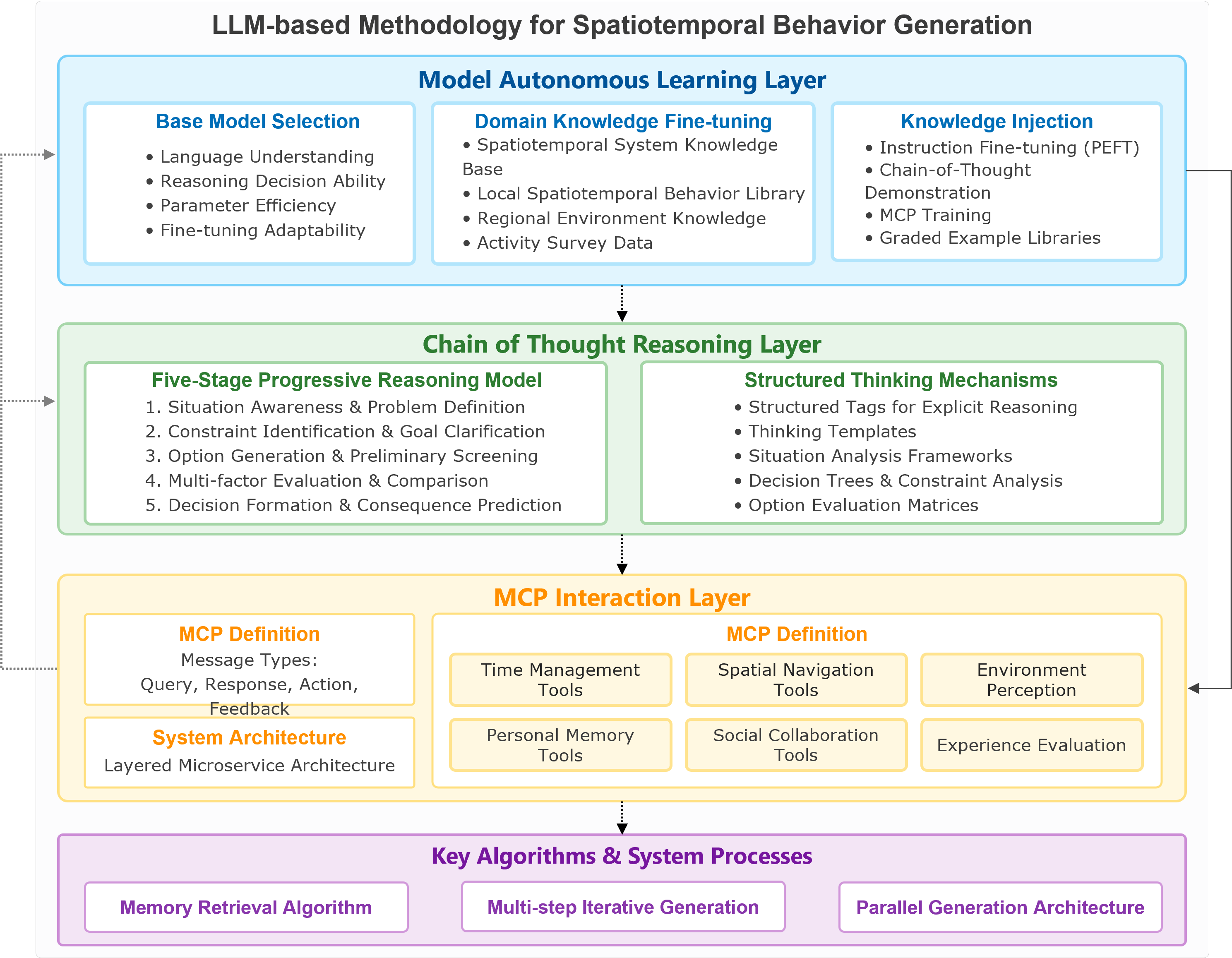}
    \end{minipage}
    \caption{LLM-based System Architecture and Methodology Framework}
    \vspace{0.5em}
    \footnotesize
    Left: Universal LLM Implementation Architecture; Right: LLM-based Methodology for Spatiotemporal Behavior Generation
    \label{fig:combined_architecture}
\end{figure}

Based on the theoretical foundation established above, this study proposes a hierarchical architecture comprising three core layers (Figure \ref{fig:combined_architecture}): the autonomous model learning layer, the CoT reasoning layer, and the MCP interaction layer. These layers form a structured hierarchy that establishes a comprehensive methodology for individual spatiotemporal behavior generation. Each layer serves distinct yet interconnected functional purposes, employing specific technical approaches that correspond to three operational mechanisms of LLMs: direct generation based on implicit knowledge, deep reasoning through explicit inference, and environmental perception and interaction capabilities through MCP that encompass both external environmental and inter-individual dynamics.

\subsubsection{Autonomous Model Learning Layer}
The autonomous model learning layer represents the fundamental mechanism by which LLMs acquire foundational knowledge and domain-specific reasoning capabilities. This layer establishes the knowledge foundation upon which all subsequent operations are based, and it comprises two principal elements: foundation model selection and domain knowledge fine-tuning.

\paragraph{(1)Foundation Model Selection} \mbox{}

Foundation model selection constitutes the critical initial phase of system development, as it directly determines the system's comprehension capabilities and reasoning performance. This study evaluates four key dimensions that ensure optimal model performance when selecting open-source foundation models, with reference to established ranking platforms such as the Hugging Face Open LLM Leaderboard: language understanding capability, reasoning and decision-making capacity, parameter efficiency, and fine-tuning adaptability.

Language understanding capability serves as the foundational requirement, demanding that the model demonstrates robust comprehension of the target language while possessing comprehensive commonsense knowledge about real-world patterns. Reasoning and decision-making capacity builds upon this foundation, requiring strong logical reasoning abilities, particularly for models capable of generating coherent chain-of-thought processes and natively supporting tool invocation. Parameter efficiency addresses deployment constraints by achieving optimal balance between computational resources and performance, thereby maximizing model capabilities while minimizing computational overhead. Finally, fine-tuning adaptability ensures rapid domain adaptation through strong few-shot learning capabilities, enabling effective adaptation to target domains with limited annotated data.
    
\paragraph{(2)Domain Knowledge Fine-tuning} \mbox{}

The domain knowledge fine-tuning process establishes two complementary knowledge repositories that enhance model training effectiveness. The first repository comprises a specialized spatiotemporal behavior knowledge base that integrates classical research findings from sociology, urban planning, and geography, organized into three components: fundamental concepts, empirical findings, and application cases. This knowledge base encompasses core theories such as time geography and activity pattern theory within its fundamental concepts component, establishing conceptual networks through definitions, explanations, and associations. The empirical findings component aggregates quantitative research discoveries from global studies, incorporating specific statistical data, behavioral metrics, and measurable relationships between environmental factors and spatiotemporal behaviors. These findings are categorized into "theoretical foundations of human behavior," "spatial and modal choice preferences," and "temporal utilization patterns", providing the model with concrete numerical evidence such as activity frequencies, travel distances, mode choice percentages, and environmental impact coefficients. The application cases component provides typical research examples that serve as specific application contexts for the model\footnote{The application cases component primarily serves future research extensions, aiming to enable individuals to provide precise decision references for urban planning from a microscopic behavioral perspective based on their spatiotemporal behavior data, thereby achieving participatory urban renewal.}.

The second repository constitutes a local spatiotemporal behavior database developed for specific research areas, documenting spatial characteristics, activity patterns, and population behaviors within those regions. This database employs multi-source data fusion methods and incorporates four data types:  \ding{172} regional environmental knowledge including POI, road networks, buildings, land use, and other spatial environmental elements;  \ding{173} authentic population activity logs collected through structured questionnaire surveys, capturing activity chain characteristics, temporal utilization patterns, travel choice behaviors, social companionship patterns, and personal subjective experiences across different demographic groups, thus reflecting individual-level spatiotemporal behavioral decision patterns and providing microscopic behavioral samples for the model;  \ding{174} spatiotemporal big data primarily derived from anonymized mobile signaling data for extracting macroscopic statistical characteristics and activity patterns of population spatiotemporal behavior;  \ding{175} social media data from domestic platforms (e.g., Xiaohongshu, Weibo) including geotagged posts, check-in records, and location-based social interactions that reveal activity preferences, place sentiments, and socio-cultural contexts of spatiotemporal behaviors.

These knowledge repositories function synergistically: the knowledge base provides universal theories and patterns, while the local database offers region-specific environmental characteristics. Together, these repositories enhance the model's activity decision-making capabilities in specific contexts.

\paragraph{(3)Knowledge Injection Methods} \mbox{}

Building upon the comprehensive knowledge repositories, three sequential knowledge injection methods are implemented to progressively enhance LLMs' capabilities. First, instruction fine-tuning employs Parameter-Efficient Fine-Tuning (PEFT) techniques to strengthen the LLMs' understanding of spatiotemporal behavior-related instructions. The instruction set encompasses diverse spatiotemporal decision tasks including activity planning, route selection, and temporal scheduling, with each instruction explicitly defining input conditions and expected outputs.

Second, CoT Demonstration Learning extracts reasoning processes from high-parameter models to construct activity decision example repositories. Each example includes complete decision steps encompassing problem analysis, constraint identification, solution generation, and evaluation selection. These examples are classified by cognitive complexity, forming progressive learning pathways that guide the LLMs from simple to complex reasoning scenarios. The inclusion of common error patterns and correction processes helps the LLMs recognize and avoid typical decision pitfalls such as constraint omission, preference misinterpretation, and spatiotemporal conflicts.

Finally, MCP Training focuses on developing the LLMs' ability to understand and follow contextual protocols. This training enables LLMs to make appropriate tool selections and execute correct usage patterns based on contextual knowledge, while maintaining protocol consistency across diverse interaction scenarios. Through this three-stage approach, LLMs progressively develop from basic instruction following to contextual reasoning capabilities.

\subsubsection{CoT Reasoning Layer}
The CoT reasoning layer employs a five-stage progressive structure that decomposes complex activity decisions into explicit reasoning steps. The method employs a structured decision-making framework inspired by cognitive science literature, with each stage building upon previous insights to facilitate computational reasoning that approximates systematic deliberation processes.

\textbf{Stage 1: Situational Awareness and Problem Definition.} Establishes foundational context through analysis of individual characteristics, temporal conditions, location, and environmental factors.

\textbf{Stage 2: Constraint Identification and Goal Clarification.} Identifies relevant temporal limitations, spatial accessibility constraints, personal preferences, and social norms while constructing multi-layered objective systems. Particular attention is given to potentially conflicting constraints and requirements that may emerge during the decision process.

\textbf{Stage 3: Option Generation and Preliminary Screening.} Leverages established situational context and identified constraints to generate diverse activity choices before conducting preliminary screening. This process evaluates option feasibility while ensuring consistency with individual behavioral patterns, creating a refined set of viable alternatives.

\textbf{Stage 4: Multi-factor Evaluation and Comparison.} Conducts comprehensive evaluation of screened options through analysis of multidimensional factors including temporal costs, spatial distance, activity value, personal preferences, and cognitive tendencies. This evaluation process aligns with authentic human cognitive characteristics, ensuring realistic reasoning patterns.

\textbf{Stage 5: Decision Formation and Consequence Prediction.} Synthesizes comprehensive evaluation results to select optimal solutions while predicting subsequent impacts on broader activity chains. This stage establishes environmental feedback mechanisms that ensure individual decisions integrate effectively into overall activity sequences and maintain adaptability to environmental changes.

The approach implements structured cognitive templates and markup systems to guide decision processing through explicit reasoning workflows encompassing situational analysis, constraint evaluation, and decision trees. Unlike traditional "black box" models, this structured chain-of-thought mechanism enhances interpretability by recording complete decision processes while incorporating human cognitive characteristics. Such integration improves both accuracy and realism of spatiotemporal behavior generation, producing decision-making patterns that authentically reflect human reasoning processes.

\subsubsection{MCP Interaction Layer}
The MCP interaction layer is designed as a layered microservice architecture to address the challenges in generating individual spatiotemporal activity-travel chains. The system design decomposes functionalities into four collaborative tiers: the Protocol Layer is designed to establish standardized message formats and communication protocols; the Service Layer comprises six core tool suites implemented as independently deployable microservices; the Integration Layer facilitates connectivity with external data sources and third-party systems; and the Optimization Layer incorporates caching mechanisms, batch processing capabilities, and load balancing strategies. The architecture employs an event-driven paradigm with inter-component communication mediated through a unified message bus infrastructure, ensuring loose coupling, high cohesion, and system extensibility for iterative functional enhancements and performance optimization.

\paragraph{(1)Protocol Definition} \mbox{}

MCP employs structured message formats comprising four fundamental message types:  \ding{172}Query messages initiated by large language models (LLMs) containing query types, parameters, and contextual information;  \ding{173}Response messages returned by MCP services including data, status indicators, and metadata;  \ding{174}Action messages representing model environmental interventions, modifications, or external tool invocations;  \ding{175}Feedback messages providing operation execution results and updated system states.

All message types utilize unified JSON encoding with clear structural and semantic definitions. MCP supports message referencing and session history maintenance, enabling models to preserve contextual coherence across multi-turn interactions.

\paragraph{(2)Core Interaction Tool Suite} \mbox{}

Six core tool categories are implemented for individual spatiotemporal activity-travel chain generation. These tools can operate independently or collaboratively to create complex behavioral sequences:

\begin{itemize}
	\item \textbf{Temporal Management Tools} (Details in ~\ref{subsec:appendix_temporal_management_tools}) process time-related queries and operations including datetime queries, interval analysis, duration estimation, and schedule conflict detection. These tools maintain unified timeline representations supporting temporal reasoning such as overlap detection and time window calculation.

	\item \textbf{Spatial Navigation Tools} (Details in ~\ref{app:Spatial Navigation Tools}) handle spatial queries and operations encompassing point-of-interest (POI) searches, route planning, and regional characterization. The system constructs bidirectional semantic-coordinate mappings and supports multi-modal travel planning for walking, public transit, driving, and cycling modes.

	\item \textbf{Environmental Perception Tools} (Details in ~\ref{app:Environmental Perception Tools}) provide environmental condition queries including weather, special events, and emergency situations. These tools support historical, real-time, and predictive temporal modes, enabling environmental factor consideration in activity selection from macro-meteorological to micro-venue conditions.

	\item \textbf{Personal Memory Tools} (Details in ~\ref{app:Personal Memory Tools}) manage individual historical behaviors and preferences through hierarchical cognitive structures incorporating event-level, pattern-level, and summary-level memory structures. The system supports dynamic storage, precise retrieval, real-time updates, and natural forgetting mechanisms. Additionally, it provides personalized decision support through intelligent matching of relevant historical experiences to current contexts.

	\item \textbf{Social Collaboration Tools} simulate social relationship networks and coordinate group activities through contact management and social interaction mechanisms. This module employs graph-based representations to calculate social distance and relationship strength, thereby incorporating social dimensions into activity decision-making. \textit{(Note: Current experiments simplify the implementation by focusing on close friends' activities and mutual arrangements stored in memory.)}

	\item \textbf{Experience Evaluation Tools} assess activity scenarios, locations, and routes through scoring mechanisms for subjective feelings, objective ratings, and planning recommendations. These tools primarily focus on evaluating and rating activity venues and travel routes without involving final outcome reasoning, creating evaluation databases supporting venue analysis and urban spatial optimization.
\end{itemize}

\paragraph{(3)Extension Modules} \mbox{}

Extension modules support spatiotemporal behavioral data generation and analysis through three complementary components. Activity-Travel Chain Validation provides basic quality control for LLM-generated sequences, performing feasibility checks to identify potential semantic gaps, errors, or logical inconsistencies. Benchmark Comparative Analysis evaluates the authenticity and representativeness of validated data by comparing outputs with observational big data across spatiotemporal distributions and behavioral characteristics. Visualization and Decision Support Systems enable data exploration and analysis, providing interactive visualization capabilities that display multi-level information ranging from individual trajectories to urban activity heat maps, facilitating both technical validation and practical decision-making processes.

\subsubsection{System Architecture and Workflow}
The proposed system employs a large-scale parallel generation architecture that integrates task decomposition, dynamic scheduling, and context caching to achieve computational efficiency for spatiotemporal behavior generation. The system decomposes generation tasks into hierarchical structures and employs dynamic resource allocation to balance computational loads across multiple processing units. Batch processing optimization consolidates similar inference requests to maximize throughput, while hierarchical context caching reduces redundant computations by sharing relevant information across different processing levels. The architecture is implemented using the Ray distributed computing platform in conjunction with vLLM's inference optimization techniques, enabling elastic scaling and dynamic resource allocation.

Within this architecture, the system implements a multi-step iterative workflow for spatiotemporal behavior generation that integrates the system components into a cohesive structure comprising six principal phases: \ding{172} persona configuration and contextual initialization, \ding{173} daily activity structure planning, \ding{174} activity detail specification, \ding{175} travel route optimization, \ding{176} state updating with dynamic adjustment, and \ding{177} trajectory validation and refinement. The workflow leverages CoT reasoning to evaluate constraints and alternatives, utilizing MCP for information acquisition and environmental state updates, and employing memory mechanisms to capture cumulative activity effects.

The system incorporates several concrete optimization strategies that operate through the MCP-LLM integration. The system implements adaptive temperature scheduling, employing a higher sampling temperature ($\tau = 0.7$) during chain-of-thought generation phases to encourage exploratory reasoning, while applying a lower temperature ($\tau = 0.1$) during MCP query formulation and final output generation to ensure precision and consistency. The generation process follows a hierarchical planning strategy that adopts a coarse-to-fine approach, initially establishing high-level activity structures with temporal allocations, followed by progressive refinement of specific locations and routes, thereby reducing combinatorial complexity. Additionally, the system leverages historical memory patterns and preferences to introduce soft biases in decision-making processes under similar contextual conditions.

The approach integrates three layers for LLM-based spatiotemporal behavior generation: autonomous learning for knowledge foundations, CoT reasoning for decision-making, and MCP interaction for environmental engagement. The system enables large-scale parallel generation while maintaining consistency. Experiments in a selected urban study area validate technical performance and behavioral realism.

\section{Experimental Design and Result Analysis}
\subsection{Experimental Setup}

\subsubsection{Study Area and Data}
This study examines the Lujiazui area\footnote{\textsuperscript{*}The study area is delineated by the red-bounded region shown in the figure. This research analyzes human behavioral activities and simulates detailed activity-travel chains for individuals who conduct at least one activity within this designated zone. For activities occurring within the study area, comprehensive behavioral simulations are performed. For activities outside the red-bounded region, only departure, arrival, and transit movements are simulated without detailed activity modeling. While individuals' daily activities may span the entire Shanghai metropolitan area, any person with at least one activity within the study area is included in the simulation framework, ensuring complete coverage of relevant activity-travel patterns.}\textsuperscript{*} due to its integrated urban functional system that contains diverse activities including financial services, retail commerce, and cultural recreation. The area attracts multiple population groups, including office workers, permanent residents, and visitors, thereby exhibiting diverse behavioral characteristics and varied spatiotemporal distributions. As a representative urban center in China, the research findings offer substantial applicability and research utility.

\begin{figure}[H]
    \centering
    \includegraphics[width=0.95\textwidth]{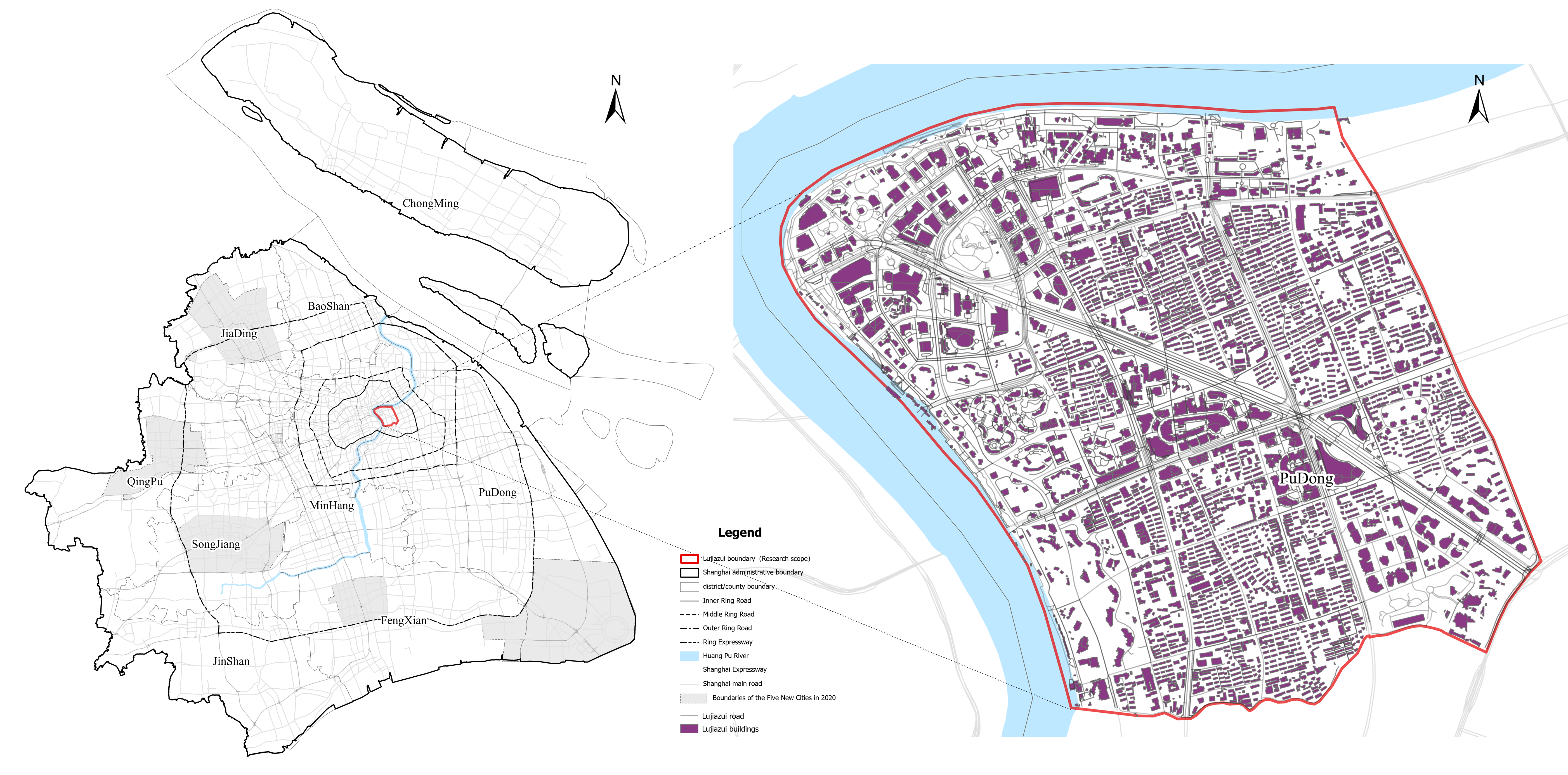}
    \caption{Delimitation map of the study area}
    \label{fig:scope}
\end{figure}

Specifically, we examine Lujiazui Subdistrict and Weifang New Village Subdistrict, bounded by Yuanshen Road to the east, Eshan Road to the south, and the Huangpu River to the northwest, covering a total area of 9.87 square kilometers (Figure\ref{fig:scope}). The delineated area includes typical urban zones such as the Financial and Trade Core Zone, Central Green Space, Century Avenue corridor, and waterfront public spaces.

The experimental data consist of:

\textbf{1. Spatiotemporal Behavioral Knowledge Base}:  A structured database containing 1,845 knowledge entries that document human spatiotemporal behavioral patterns, which were synthesized from over 1,000 peer-reviewed publications and empirical research findings across interdisciplinary fields including time geography, behavioral geography, and urban planning (see Appendix \ref{appendix:knowledge-base-architecture} for details). Each knowledge entry follows a question-answer format, averages 600 words, and addresses human behavioral patterns and decision-making mechanisms across different geographical contexts, sociodemographic groups, activity types, temporal cycles, and spatial environments.

\textbf{2.Basic Geographic Information}: This dataset includes POI data, building and road network data, and transportation network information. The POI data were cross-validated and enhanced with information from platforms such as Dianping, including detailed attributes such as establishment names, categories, addresses, coordinates, operating hours, ratings, and price levels.

\textbf{3.Spatiotemporal Behavioral Data}:  \ding{172}Activity diary survey data consisting of 571 daily activity records from diverse populations in the Lujiazui area, serving as baseline data for training dataset construction and model validation;  \ding{173}Anonymized mobile signaling data containing spatiotemporal trajectory points of approximately 150,000 users from November 2023, which underwent data cleaning and privacy protection procedures, and were utilized for aggregate-level validation of generated data by comparing spatiotemporal distribution patterns and activity characteristics to assess trend consistency;  \ding{174}Synthetic training data: Using the two aforementioned datasets, 2,000 individual activity-travel chains with "a priori rationality"\footnote{\textsuperscript{*}A priori rationality indicates that the synthetic activity-travel chains are generated with predefined statistical constraints derived from the empirical patterns observed in the activity diary and mobile signaling datasets, ensuring the generated data conform to realistic human spatiotemporal behavioral regularities rather than random sequences.}\textsuperscript{*} were generated through time-varying Markov chain modeling methods for model training and fine-tuning optimization.

\subsubsection{Experimental Design}

This study adopts a comparative experimental framework to evaluate the performance of the proposed methods in activity-travel chain generation tasks. The experimental design comprises four primary components:
\begin{itemize}
	\item \textbf{1.Foundation Model Comparison Experiment}: This experiment evaluates the performance of different foundation models in individual spatiotemporal behavior generation tasks, including GLM-4-9B-0414, DeepSeek-R1-Distill-Qwen-7B, LLAMA-2-13B, and Bloom-7B1. Evaluation dimensions include generation quality, reasoning capability, generation diversity, and computational efficiency.

	\item \textbf{2.Method Component Ablation Experiment}: To systematically evaluate the contribution of each core component to generation quality, three controlled ablation experiments are conducted, targeting implicit knowledge training, chain-of-thought reasoning, and MCP interaction respectively. The implicit knowledge training ablation includes two variants: removing the spatiotemporal behavioral professional knowledge base and removing study area activity data; the chain-of-thought reasoning ablation replaces the five-stage progressive reasoning model with direct generation; the MCP interaction tool ablation employs a sequential removal strategy to systematically evaluate six core tools.

	\item \textbf{3.Parallel Processing Scalability Experiments}: This experiment evaluates computational efficiency and scalability using parallel processing configurations. The study examines performance across different numbers of parallel workers (2, 4, 8, and 12 workers) while generating 1,000 person-day samples. System performance is measured through two key metrics: average generation time per sample and GPU memory consumption.
	
	\item \textbf{4.Benchmark Method Comparison Experiment}(to be conducted)\footnote{\textsuperscript{*}Due to time constraints and resource limitations, this comparative analysis has not yet been completed but will be pursued in subsequent research phases.}\textsuperscript{*}: A detailed comparative evaluation will benchmark the proposed method against four established baseline approaches:  \ding{172}LSTM-based activity sequence generation model;  \ding{173}Time-Varying Markov Chain model (TVMC);  \ding{174}Rule-driven activity scheduling system (utilizing utility maximization discrete choice modeling);  \ding{175}Agent-based urban activity simulation framework (ActivitySim). Assessment dimensions will include generation quality, computational efficiency, and resource overhead.
\end{itemize}

\begin{figure}[htbp]
    \centering
    \includegraphics[width=0.95\textwidth]{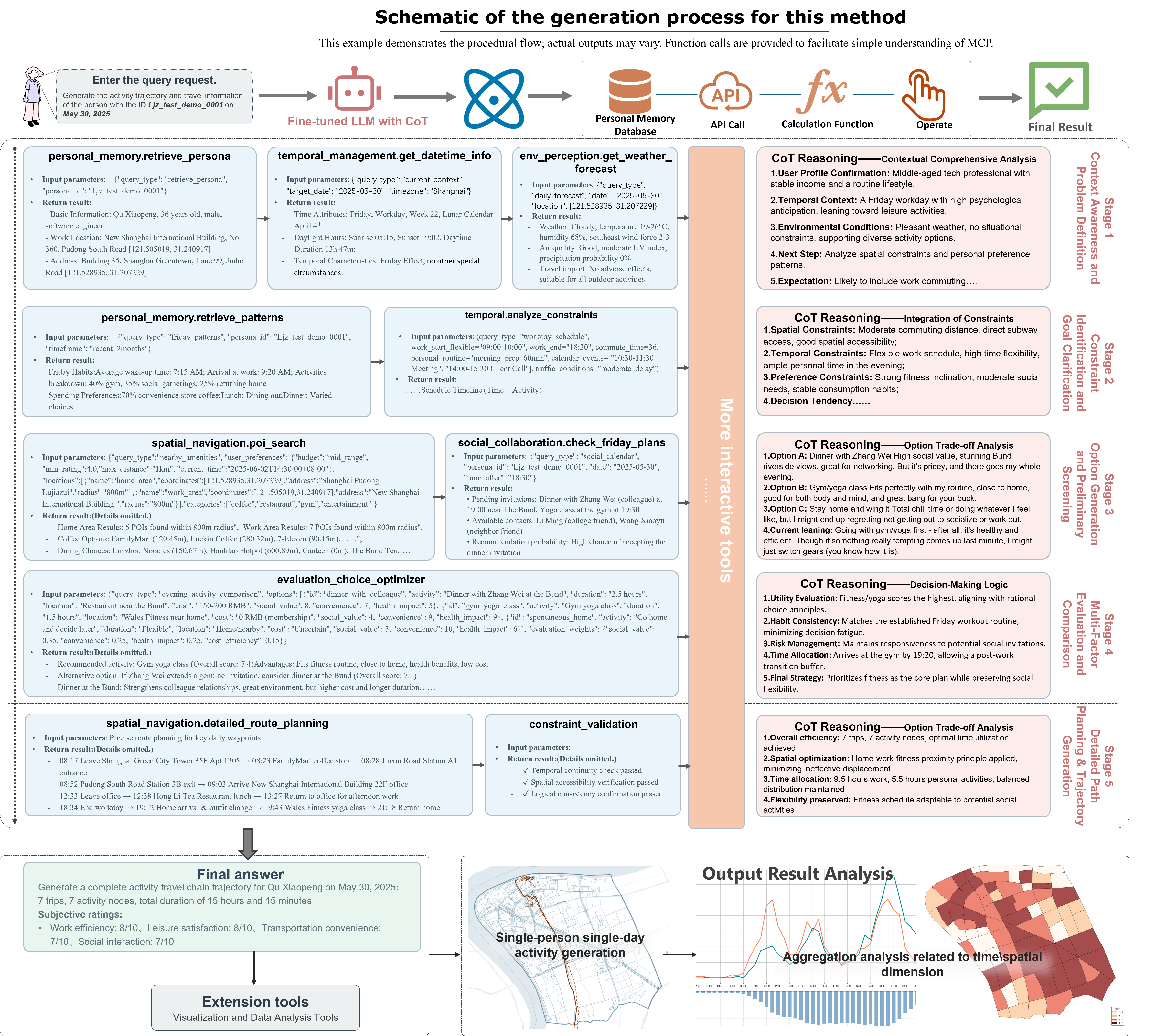}
    \caption{Schematic of the generation process for this method}
    \label{fig:fig}
\end{figure}

\subsubsection{Evaluation Method}
Due to the complexity of spatiotemporal behavior generation and evaluation, this study employs a comprehensive evaluation approach to assess the proposed methodology across four critical dimensions: generation quality, reasoning capability, generation diversity and computational performance. The evaluation adopts a multi-batch generation strategy, with each batch independently evaluated to obtain individual performance scores. The highest and lowest scoring batches are then excluded, and final metrics are calculated as the mean of the remaining batch scores. All performance metrics are standardized to a 0-10 scale to ensure comparability and consistency across different evaluation criteria. To maintain objectivity and consistency in the assessment process, \textit{GPT-4o-mini} serves as an independent third-party evaluation model, utilizing structured template-based prompts to replace conventional expert-based scoring methodologies.

\textbf{(1)Generation Quality:}

A generation quality assessment method is developed that combines subjective evaluation of individual activity patterns with objective validation using aggregate spatiotemporal data. The subjective evaluation systematically assesses generated activity-travel chains across multiple dimensions using predefined evaluation criteria. The evaluation examines four key dimensions(see \ref{app:Subjective Evaluation Criteria for Generation Quality}):  \ding{172}Temporal consistency, which examines the logical sequencing of activities, realistic nature of time allocation, and reasonable duration of transition times;  \ding{173}Activity coherence, which evaluates the logical relationships within activity sequences, alignment between trip purposes and activity types, and logical flow of activity chains;  \ding{174}Persona conformity, which verifies the alignment between generated activity patterns and specified individual characteristics;  \ding{175}Behavioral authenticity, which assesses the realistic nature of activity content, duration, and locations. Each dimension is scored on a 10-point scale, with the overall score derived through weighted averaging.

Complementing the subjective assessment, objective validation establishes quantitative benchmarks using real-world mobile signaling data. The validation conducts quantitative analysis across spatial and temporal dimensions. Spatial validation employs kernel density estimation to assess the similarity between observed and generated urban activity distributions, conducted at two temporal scales: all-day activity hotspot distribution and time-segmented hotspot distribution. The former aggregates total visit frequencies across all activity points throughout the day, while the latter analyzes time-specific population distributions at specific time intervals. Spatial distribution similarity is quantified using Jensen-Shannon divergence:

\begin{equation}
    JS(P \parallel Q) = \frac{1}{2} D_{\text{KL}}(P \parallel M) + \frac{1}{2} D_{\text{KL}}(Q \parallel M),
\end{equation}
where $ P $ and $ Q $ represent probability distributions of real and generated data respectively, and $ M = \frac{1}{2}(P + Q) $. The JS divergence ranges from 0 to 1, with smaller values indicating higher distributional similarity.

Temporal validation focuses on the distributional similarity of start times, end times, and durations across various activity types, employing the Kolmogorov-Smirnov test:

\begin{equation}
    D_{\text{KS}} = \sup_{x} \big| F_n(x) - F_m(x) \big|,
\end{equation}
where $ F_n(x) $ and $ F_m(x) $ denote the empirical cumulative distribution functions (CDFs) of real and generated data, respectively. Furthermore, daily temporal distribution patterns are validated by comparing population presence probability density curves across different time periods.

The overall objective quality score integrates spatiotemporal indicators:
\begin{equation}
    Q_{\text{obj}} = 0.5 \times \left(1 - \frac{\text{JS}_{\text{spatial}}}{\ln(2)}\right) + 0.5 \times (1 - \text{KS}_{\text{temporal}})
\end{equation}
The final generation quality is computed as:
\begin{equation}
    Q_{\text{total}} = \alpha \times Q_{\text{subjective}} + (1 - \alpha) \times Q_{\text{obj}},
\end{equation}
where $\alpha$ represents the weighting parameter that can be adjusted according to application scenarios to balance the relative importance of subjective and objective evaluations (set to $\alpha = 0.5$ in this study).

\textbf{(2)Reasoning Capabilities:}

 Reasoning capabilities are assessed by analyzing CoT quality during spatiotemporal behavior generation. Other large language models are employed to evaluate reasoning chains across five critical dimensions(see \ref{Reasoning Capabilities Evaluation Criteria} for details).

 Information collection completeness examines whether models identify temporal constraints, spatial positioning, user preferences, resource limitations, environmental factors, etc. Analytical thinking depth evaluates analysis across temporal, spatial, cost-benefit, risk assessment, and personalization considerations. Logical reasoning rigor examines causal relationships between reasoning steps, assumption validity, and the logical foundations of conclusions. Decision-making rationality assesses how transparently models balance conflicting constraints. Tool invocation strategy analyzes the appropriateness of timing, sequencing, and parameter configuration.

\textbf{(3)Generation Diversity:}

The DTW-clustering sequence pattern-based approach for assessing generative diversity systematically transforms continuous spatiotemporal behaviors into discrete sequential patterns, thereby enabling precise quantification of sequence similarity through Dynamic Time Warping (DTW) distance metrics. This approach constructs activity sequences of length 96 with 15-minute temporal granularity, represented as $\mathbf{S} = \{s_1, s_2, \ldots, s_{96}\}$, where each element $s_t \in \{\text{life services}, \text{sports \& leisure}, \text{employment}, \text{tourism}, \text{residence}, \text{shopping}, \text{dining}, \text{travel}\}$ denotes the predominant activity category at time step $t$.

The DTW algorithm serves as the foundation for measuring sequence similarity by constructing a cumulative distance matrix and employing dynamic programming to identify the optimal alignment path. The recurrence relation is formulated as:

\begin{equation}
\text{DTW}(i,j) = d(s_i, s_j) + \min \begin{cases} 
\text{DTW}(i-1,j) & \text{(insertion)} \\ 
\text{DTW}(i,j-1) & \text{(deletion)} \\ 
\text{DTW}(i-1,j-1) & \text{(match)}
\end{cases}
\end{equation}

where $d(s_i, s_j)$ represents the local distance function between sequence elements $s_i$ and $s_j$. For computational efficiency, the distance function employs a binary measure: $d(s_i, s_j) = 0$ if $s_i = s_j$, and $d(s_i, s_j) = 1$ otherwise (treating all inter-category distances as equivalent despite their varying semantic relationships).

Subsequently, for the ensemble of $N$ generated activity chains, pairwise DTW distances are computed to construct an $N \times N$ dissimilarity matrix $\mathbf{D}$. To identify underlying behavioral patterns within this high-dimensional space, hierarchical clustering is applied using Ward's linkage criterion, which minimizes the increment in within-cluster sum of squares. The merge distance between clusters $C_i$ and $C_j$ is defined as:

\begin{equation}
d(C_i, C_j) = \sqrt{\frac{2n_i n_j}{n_i + n_j}} \cdot \|\boldsymbol{\mu}_i - \boldsymbol{\mu}_j\|_2
\end{equation}

where $n_i$ and $n_j$ denote the number of samples in clusters $C_i$ and $C_j$, respectively, and $\boldsymbol{\mu}_i$ and $\boldsymbol{\mu}_j$ represent their corresponding cluster centroids in the DTW distance space.

The optimal number of clusters $k^*$ is determined by maximizing the silhouette coefficient:

\begin{equation}
s_i = \frac{b_i - a_i}{\max\{a_i, b_i\}}
\end{equation}

where $a_i$ represents the mean intra-cluster distance for observation $i$, and $b_i$ denotes the mean distance to the nearest neighboring cluster. The silhouette coefficient ranges from -1 to 1, with higher values indicating better clustering quality.

The diversity assessment employs between-cluster average distance $B$ and within-cluster average distance $W$ to construct the diversity index:

\begin{equation}
\text{DIV} = \frac{B - W}{B + W} \quad \text{where } \text{DIV} \in [-1, 1]
\end{equation}

This index is transformed to a standardized scoring scale using:

\begin{equation}
\text{Score}_{\text{GD}} = 5.5 + 4.5 \times \text{DIV} \quad \text{where } \text{Score}_{\text{GD}} \in [1, 10]
\end{equation}

where $\text{Score}_{\text{GD}}$ represents the Generation Diversity score, with higher values indicating greater pattern diversity across activity chain sets.

\textbf{(4)Computational Performance:}

Computational performance was evaluated across two dimensions: computational efficiency and resource consumption. All experiments were conducted on a standardized testing environment\footnote{\textsuperscript{*}Testing environment specifications: NVIDIA A800-80GB GPU, Intel Xeon Gold 6348 processor, 100GB RAM, CUDA 12.4 runtime environment. The results may vary depending on different deployment environments and hardware configurations.}\textsuperscript{*}.

Computational efficiency was quantified using average generation time as the primary metric, representing the time required for the system to generate a complete daily spatiotemporal behavior pattern for a single individual, expressed in seconds per person-day (s/person·day). The standard deviation of generation times was calculated to assess performance stability. Resource consumption analysis focused on GPU memory utilization during system operation, with real-time monitoring capturing both average and peak memory usage to identify baseline requirements and potential bottlenecks.

\subsection{Experimental Results}

\subsubsection{Base Model Comparison Results}

\begin{figure}[htbp]
    \centering
    \includegraphics[width=0.8\textwidth]{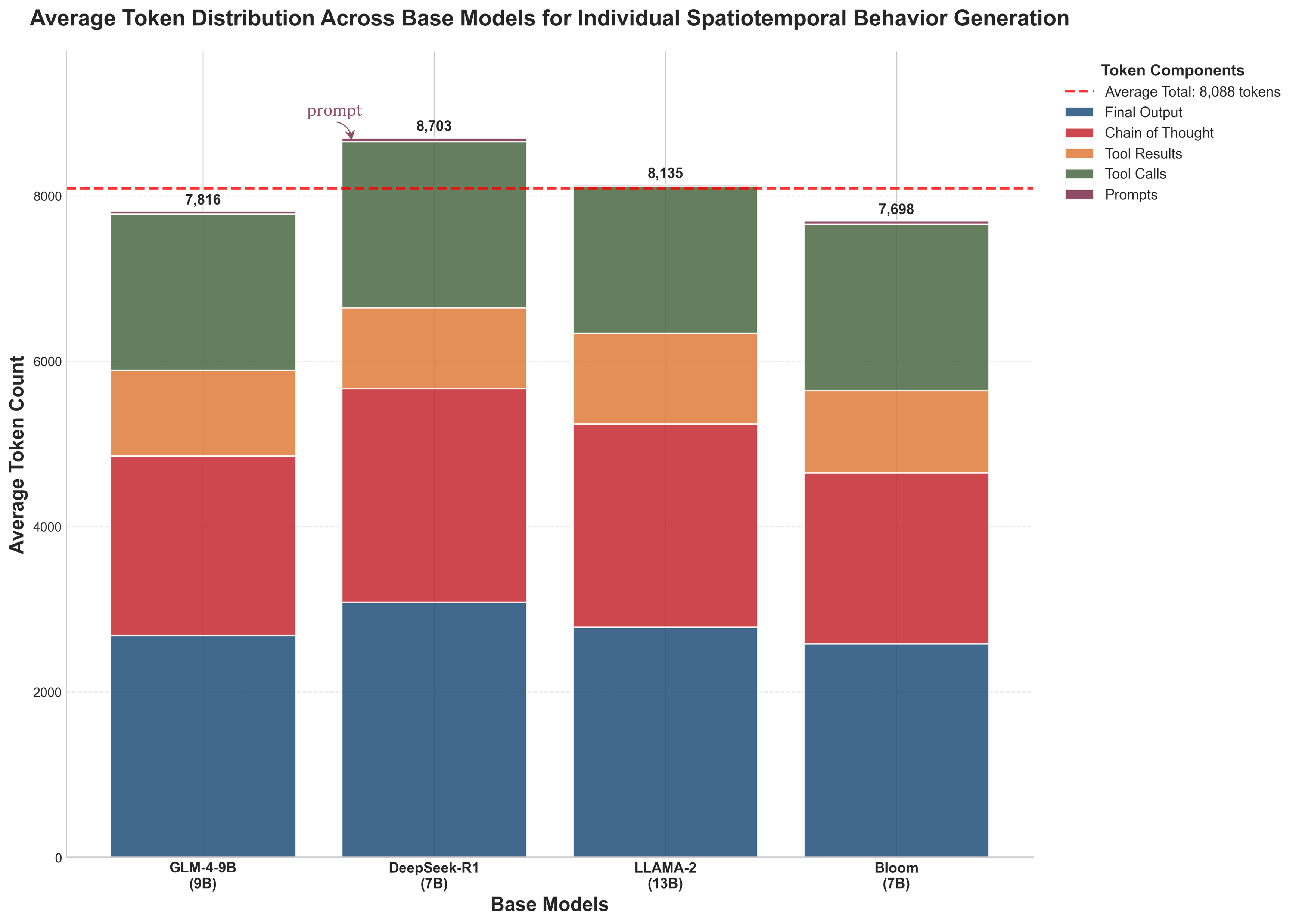}
    \caption{Average Token Distribution Across Base Models for Individual Spatiotemporal Behavior Generation}
    \label{fig:Average_tokens}
\end{figure}

We conducted this experiment using the proposed method under standardized conditions. To assess performance differences among base models, four selected open-source base models were evaluated. Given that parameter scale affects performance, the selected models covered a parameter range from 7B to 13B parameters. The experimental design comprised 10 batches of 50 samples for each model, yielding 500 samples per model and a total dataset of 2,000 samples. Each generation process for a single data instance involved an average of 12-18 tool calling sequences, encompassing geographic information queries, temporal planning, activity reasoning, path optimization, and other computational steps, with an average token consumption of approximately 8,000 tokens(Figure \ref{fig:Average_tokens}). The experimental results indicate that all four models perform comparably across all dimensions and can successfully complete spatiotemporal behavior generation tasks, but nevertheless exhibit their own characteristics in subtle differences and specific scenarios.

The experimental results(Table \ref{tab:performance_comparison}) show consistent performance across different base models under the MCP+CoT framework, despite significant differences in parameter count and architecture. Average generation quality scores differ by only 0.50 points across all models. Notably, the 7B DeepSeek model outperforms the 13B LLAMA-2 model (8.36 vs. 8.22), indicating that parameter count does not determine performance quality. This consistency extends to other evaluation dimensions: reasoning capabilities remain similar across models (CoT scores: 7.24-7.63), and computational efficiency shows stable performance (generation time: 2.2-3.0 min/person-day) regardless of model architecture. These findings suggest that the MCP+CoT framework supports relatively consistent task performance, Seemingly independent of the underlying model complexity under specific test configurations and task constraints. Success appears to depend primarily on the model's basic world knowledge and ability to use tools correctly, rather than architectural sophistication. The differences between models reflect implementation details rather than fundamental capability gaps, confirming that the framework works effectively across diverse model configurations.

Given the comparable performance across all models, GLM-4-9B was selected for subsequent experiments primarily due to its specialized optimization for Chinese language processing and superior contextual understanding of urban spatiotemporal behaviors. While DeepSeek-R1-7B achieved the highest generation quality score (8.36), GLM-4-9B demonstrates better semantic comprehension and cultural adaptability specifically for Chinese contexts, which is critical for generating realistic spatiotemporal behavior patterns. Its deep training on Chinese corpora enables more accurate interpretation of location semantics, activity patterns, and cultural behavioral norms essential for this task domain.

\renewcommand{\arraystretch}{1.5}
\setlength{\extrarowheight}{2pt}
\setlength{\arrayrulewidth}{0.2pt}
\begin{longtable}{>{\centering\arraybackslash}m{2.1cm}>{\centering\arraybackslash}m{2.5cm}>{\centering\arraybackslash}m{2.2cm}>{\centering\arraybackslash}m{2.5cm}>{\centering\arraybackslash}m{2.2cm}>{\centering\arraybackslash}m{2cm}}
\caption{Performance Comparison of Base Models for Individual Spatiotemporal Behavior Generation (N = 4 $\times$ 500, n = 10 $\times$ 50)} \\
\label{tab:performance_comparison} \\
\toprule
\textbf{Evaluation Dimension} & \textbf{Sub-dimension/Metric} & \textbf{GLM-4-9B-0414 (9B)} & \textbf{DeepSeek-R1-Distill-Qwen-7B (7B)} & \textbf{LLAMA-2-13B (13B)} & \textbf{Bloom-7B1 (7B)} \\
\midrule
\endfirsthead
\toprule
\textbf{Evaluation Dimension} & \textbf{Sub-dimension/Metric} & \textbf{GLM-4-9B-0414 (9B)} & \textbf{DeepSeek-R1-Distill-Qwen-7B (7B)} & \textbf{LLAMA-2-13B (13B)} & \textbf{Bloom-7B1 (7B)} \\
\midrule
\endhead
\multirow{3}{*}{\parbox{2cm}{\textbf{Generation Quality}}} & Subjective Individual ($Q_{\text{subjective}}$) & 7.45 $\pm$ 1.07 & 7.61 $\pm$ 0.80 & 7.54 $\pm$ 1.03 & 7.22 $\pm$ 0.91 \\
& Objective Aggregate ($Q_{\text{obj}}$) & 8.70 $\pm$ 0.45 & 9.10 $\pm$ 0.42 & 8.90 $\pm$ 0.38 & 8.50 $\pm$ 0.48 \\
& \textbf{Weighted Average ($Q_{\text{total}}$)} & \textbf{8.07 $\pm$ 0.58} & \textbf{8.36 $\pm$ 0.45} & \textbf{8.22 $\pm$ 0.55} & \textbf{7.86 $\pm$ 0.51} \\
\midrule
\multirow{1}{*}{\parbox{2cm}{\textbf{Reasoning Capability}}} & \textbf{CoT Quality Assessment} & \textbf{7.55 $\pm$ 1.11} & \textbf{7.62 $\pm$ 0.78} & \textbf{7.63 $\pm$ 1.05} & \textbf{7.24 $\pm$ 0.83} \\
\midrule
\multirow{2}{*}{\parbox{2cm}{\textbf{Generation Diversity}}} & Behavior Pattern Clusters ($k^*$) & 11 & 9 & 15 & 8 \\
& \textbf{Diversity Score} & \textbf{7.28} & \textbf{7.19} & \textbf{7.71} & \textbf{6.83} \\
\midrule
\multirow{3}{*}{\parbox{2cm}{\textbf{Computational Performance}}} & Generation Time (min/person-day) & 2.4 $\pm$ 0.9 & 2.7 $\pm$ 0.7 & 3.0 $\pm$ 1.1 & 2.2 $\pm$ 0.5 \\
& GPU Memory Usage (GB) & 19.8 $\pm$ 2.5 & 16.6 $\pm$ 2.2 & 23.9 $\pm$ 3.2 & 16.2 $\pm$ 1.7 \\
& Peak Memory Usage (GB) & 22.8 & 19.5 & 28.7 & 19.6 \\
\bottomrule
\end{longtable}
% 重置行间距为默认值
\renewcommand{\arraystretch}{1}
\setlength{\extrarowheight}{0pt}

\subsubsection{Component Ablation Study Results}

\begin{figure}[htbp]
    \centering
    \includegraphics[width=0.8\textwidth]{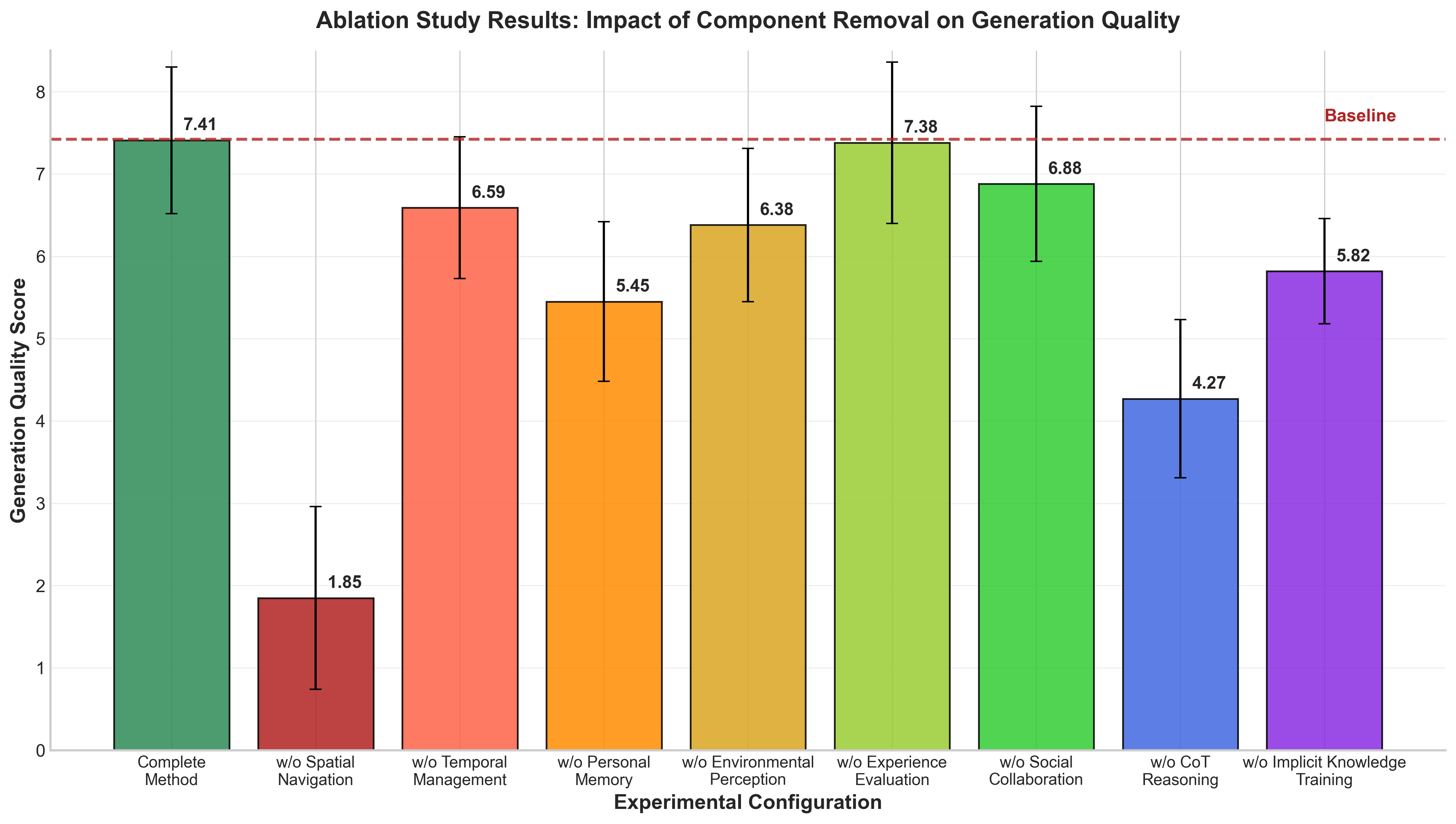}
    \caption{Ablation Study Results: Impact of Component Removal on Generation Quality}
    \label{fig:fig_Ablation}
\end{figure}

To systematically assess the differential contributions of each component to spatiotemporal behavior generation quality, ablation experiments were performed with GLM-4-9B as the baseline model. Individual components were sequentially removed to quantify their impact on system performance (n=500). The experimental results are shown in Figure \ref{fig:fig_Ablation}.

The ablation study reveals a clear hierarchy of component importance, with the spatial navigation tool serving as the foundational element of the entire system. Its removal resulted in severe performance degradation, reducing generation quality to 1.85 points—a 75.1\% decline from baseline performance. This substantial degradation primarily stems from the system's inability to process spatial location information and maintain geographic constraints. The high standard deviation reflects a bimodal distribution of output quality: generated samples were either completely unusable or marginally acceptable, with the latter primarily relying on recombination of knowledge injected during the fine-tuning stage rather than systematic spatial reasoning.

CoT reasoning demonstrated the second-highest impact on system performance, with its absence causing a substantial 42.4\% degradation to 4.27 points. Without this logical scaffolding, the system regressed to direct generation based on statistical patterns, losing its capacity for stepwise reasoning and multi-factor consideration. This degradation was particularly pronounced in complex decision scenarios, where the decision-making process became completely opaque and generated outputs exhibited excessive homogenization, resembling results from probability distribution fitting rather than reasoned human behavior. The substantial standard deviation indicates varying dependency on reasoning mechanisms across different task types.

Personal memory and implicit knowledge training both contributed significantly to system capabilities. The personal memory tool's removal led to a 26.5\% performance decline (5.45 points), primarily manifesting as severe homogenization of individual behavior patterns and complete loss of learning adaptation capabilities. DTW clustering analysis revealed that behavior pattern clusters dramatically reduced from 11 to 5, confirming substantial diversity loss. Similarly, eliminating implicit knowledge training resulted in a 21.5\% degradation (5.82 points), as the system lost its foundation of domain expertise and social normativity, making generated behaviors deviate significantly from realistic human behavioral patterns.

Other components demonstrated varying degrees of impact on system performance. Environmental perception tool absence caused a 13.9\% decline (6.38 points), primarily affecting the system's ability to perceive external environmental factors and adapt contextually to real-world conditions. Temporal management tool removal resulted in a 11.0\% degradation (6.59 points), with impacts mainly on the system's understanding of daily activity timelines and temporal flow coordination. Social collaboration tool showed minimal effects (7.15\% degradation to 6.88 points), which may be attributed to the current implementation's limited scope (primarily adding friends' activity patterns to personal memory or generating social invitations, rather than comprehensive interpersonal social interaction processing) and the experimental dataset's focus on individual activities. The experience evaluation tool's removal showed negligible impact with only a slight decline to 7.38 points, consistent with its designed role in supporting future research extensions rather than core generation logic.

The ablation results reveal distinct importance levels across system components. Chain-of-thought reasoning appears essential for maintaining logical coherence in the tested scenarios, with its removal resulting in substantial performance degradation under the experimental conditions. Within the MCP tool system, spatial navigation emerges as the foundational component, with its absence leading to near-complete system failure. Other MCP tools show varying importance: personal memory and environmental perception contribute meaningfully to performance, while social collaboration and experience evaluation demonstrate minimal impact. These findings suggest that CoT reasoning and spatial navigation should be considered essential components, while other MCP tools can be configured selectively based on application requirements and computational constraints.

\subsubsection{Parallel Processing Scalability Experiments}

To evaluate the computational efficiency and scalability of the MCP CoT methodology, parallel performance evaluations were conducted using the unquantized GLM-4-9B model on A800 80G hardware. The experiment generated 1,000 person-day samples while varying the number of parallel workers from 2 to 12. Each worker operates independently to process individual sample generation tasks, enabling simultaneous generation of multiple person-day data samples without task interdependence. All experiments utilized the vLLM inference optimization engine.

\begin{table}[htbp]
\centering
\caption{Performance Metrics under Different Parallel Configurations}
\label{tab:parallel_performance}
\renewcommand{\arraystretch}{1.2}
\begin{tabular}{ccc}
\hline
\textbf{Parallel Workers} & \textbf{Mean Generation Time} & \textbf{Mean Memory Usage} \\
\textbf{} & \textbf{(minutes/sample)} & \textbf{(GB)} \\
\hline
2 & 1.30 & 38.4 \\
4 & 0.70 & 51.7 \\
8 & 0.30 & 67.2 \\
12 & 0.17 & 76.8 \\
\hline
\end{tabular}
\end{table}

Increasing from 2 to 8 parallel workers achieved substantial performance improvements, with mean generation time per sample decreasing from 1.30 minutes to 0.30 minutes, representing a 77\% reduction in processing time. Memory consumption scaled predictably from 38.4GB to 67.2GB, remaining within the 80GB system capacity. Scaling to 12 workers further reduced generation time to 0.17 minutes while consuming 76.8GB of memory. However, acceleration gains diminish at higher parallelization levels due to memory bandwidth constraints and inter-worker communication overhead.

The results demonstrate that increasing parallelization achieves substantial performance improvements through independent task distribution across multiple workers. This validates that the MCP CoT methodology can effectively increase generation speed by scaling parallel workers to meet large-scale sample generation requirements, though efficiency gains diminish at higher worker counts due to hardware bottlenecks.

\section{Discussion}

\subsection{Main Findings and Contributions}

This investigation establishes an MCP-enhanced chain-of-thought framework for urban spatiotemporal behavior generation, contributing to theoretical, technical, and practical advances that partially address persistent challenges in computational urban behavior modeling.

\begin{itemize}
	\item Theoretical Innovation. This research develops a novel integration of chain-of-thought reasoning with Model Context Protocol to create interpretable simulation frameworks for human activity decision-making. The system connects natural language understanding capabilities with urban computing tools, thereby bridging cognitive reasoning processes and spatiotemporal modeling requirements.
	\item Technical Implementation. The framework employs a comprehensive MCP interaction system with specialized tool categories and layered microservice architecture designed for large-scale parallel urban spatiotemporal behavior generation.
	\item Application Validation. Empirical validation in Shanghai's Lujiazui district demonstrates the framework's capacity to generate realistic activity patterns that reflect authentic urban behavior. The approach enables zero-shot deployment without additional training data, indicating notable potential for scalable applications in urban planning and transportation management.
\end{itemize}

\subsection{Method Limitations}
Despite these substantial advances, several interconnected limitations constrain the methodology's current scope and effectiveness, necessitating acknowledgment and future development efforts.

\begin{itemize}
	\item Evaluation Framework Complexities. The evaluation of individual spatiotemporal behavior generation presents considerably greater complexity than conventional large language model assessment tasks. Activity chains encompass intricate individual decision-making processes whose reasonableness and authenticity cannot be adequately quantified through conventional text similarity or accuracy metrics. Current evaluation systems rely predominantly on distributional similarity comparisons between generated outputs and empirical data. However, such macroscopic statistical matching approaches cannot effectively assess microscopic behavioral reasonableness or the internal consistency of decision logic. Furthermore, human behavior exhibits inherent subjectivity and uncertainty, precluding the establishment of definitive ground truth evaluation benchmarks.
	\item Individual Behavior Modeling Complexity. The complexity of human spatiotemporal behavior extends beyond the scope of linguistic expression, involving dynamic interactions among cognitive, emotional, social, and environmental factors. Individual behavior frequently contains unpredictable elements of randomness and innovation that remain challenging to anticipate accurately. While the proposed methodology enhances decision process reasonableness and transparency through chain-of-thought reasoning mechanisms, limitations persist in capturing individual uniqueness, emotional fluctuations, and abrupt preference modifications. Moreover, the underlying interpretability of decision-making processes remains constrained by the token-based generation characteristics of LLMs, wherein fundamental reasoning mechanisms remain opaque.
	\item Spatiotemporal Scale and Generalization Limitations. Current experimental validation is primarily confined to Lujiazui, a specific urban central district, limiting empirical assessment of cross-regional generalization capabilities and adaptability to diverse urban morphologies and cultural contexts. Temporally, the methodology focuses on daily activity chain generation with testing limited to one typical workday, requiring deeper investigation into behavioral evolution patterns across extended temporal scales. While the framework possesses theoretical capacity for zero-shot transfer to novel study areas, empirical validation of both spatial transferability and temporal scalability remains constrained by resource and time limitations, leaving cross-regional and long-term behavioral modeling capabilities empirically unvalidated.
\end{itemize}

\subsection{Future Research Directions}

Based on current achievements and identified limitations, future research should focus on advancing the following key aspects:

\begin{itemize}
	\item Multi-dimensional Evaluation Framework Challenges: This presents a fundamental paradox representing one of the most challenging aspects in this field: if definitive evaluation methods existed to accurately assess synthetic behavioral authenticity, sophisticated simulation techniques would be largely unnecessary. Current approaches validate macro-level consistency with real-world trends, yet this creates potential vulnerability to overfitting observable data distributions without capturing underlying behavioral authenticity. Future research must develop frameworks that balance assessment rigor with practical feasibility, focusing on emergent behavioral properties rather than direct pattern matching.
	\item Multimodal Fusion and Real-time Adaptation: Integrate multimodal data sources including text, images, audio, and physiological signals to construct comprehensive individual behavior understanding models. Develop real-time environmental perception and dynamic adjustment mechanisms.
	\item Cross-temporal and Cross-spatial Validation: To evaluate the zero-shot transfer capability of this method, it is tested across different time spans and urban areas, with a particular focus on its generalization performance in diverse urban environments and cultural contexts.
	\item Social Decision-making and Collective Behavior: Model complex social interactions and group dynamics to understand how individual behaviors aggregate into collective phenomena.
	\item Anomalous Behavior and Special Population Modeling: Apply the proposed method to simulate exceptional circumstances and special population behaviors where real-world data collection is challenging or infeasible, enabling research into rare events, emergency situations, and atypical behavioral patterns.
	\item Application-oriented Optimization: Facilitate the translation of research outcomes into deployable practical systems for smart cities, transportation management, and commercial scenarios.
\end{itemize}

\section{Conclusion}

This study presents a framework for individual spatiotemporal behavior generation using MCP-enhanced chain-of-thought large language models. The proposed approach alleviates important limitations of traditional behavior modeling methodologies by integrating natural language understanding capabilities with spatiotemporal reasoning mechanisms.

Key findings from experimental validation demonstrate the framework's effectiveness across multiple dimensions. The Shanghai Lujiazui district case study shows that the methodology generates realistic activity patterns with quality scores ranging from 7.86 to 8.36 across different base models. Parallel processing experiments reveal substantial efficiency gains, reducing generation times from 1.30 to 0.17 minutes per sample when scaling from 2 to 12 processes. Component ablation studies confirm the critical importance of chain-of-thought reasoning and spatial navigation tools, while other MCP components contribute varying levels of performance enhancement.

The research contributes to the intersection of urban computing, artificial intelligence, and behavioral science through several innovations. The integration of structured reasoning processes with external tool interaction capabilities addresses fundamental challenges in capturing both individual decision-making logic and environmental constraints. The comprehensive MCP framework with specialized tool categories provides a systematic approach to spatiotemporal data processing, while the parallel processing architecture enables practical large-scale deployment.

Despite these advances, several limitations constrain the methodology's current scope. Evaluation frameworks for synthetic behavioral data remain complex, individual behavior modeling faces inherent unpredictability challenges, and spatial-temporal generalization requires broader empirical validation across diverse urban contexts and extended time periods.
Future research should focus on enhancing cross-regional generalization, incorporating real-time adaptation mechanisms, and expanding applications to diverse urban contexts. Priority areas include developing multimodal data fusion capabilities, advancing evaluation methodologies for synthetic behavioral authenticity, and extending the framework to model complex social interactions and collective behavior patterns.

The methodology's capacity to generate spatiotemporal behaviors that correspond with validation data while maintaining computational efficiency under tested conditions suggests promising directions for large-scale urban modeling applications. This work advances the integration of linguistic reasoning with spatial analysis, contributing to the development of responsive urban systems that can adapt to evolving population needs and support evidence-based urban planning decisions.

\clearpage
\bibliographystyle{unsrtnat}
\bibliography{references}  %%% Uncomment this line and comment out the ``thebibliography'' section below to use the external .bib file (using bibtex) .

\begin{thebibliography}{43}
\providecommand{\natexlab}[1]{#1}
\providecommand{\url}[1]{\texttt{#1}}
\expandafter\ifx\csname urlstyle\endcsname\relax
  \providecommand{\doi}[1]{doi: #1}\else
  \providecommand{\doi}{doi: \begingroup \urlstyle{rm}\Url}\fi

\bibitem[Wang et~al.(2020)Wang, Zhang, Chan, Shi, Zhou, and Liu]{wang2020review}
Anqi Wang, Anshu Zhang, Edwin~HW Chan, Wenzhong Shi, Xiaolin Zhou, and Zhewei Liu.
\newblock A review of human mobility research based on big data and its implication for smart city development.
\newblock \emph{ISPRS International Journal of Geo-Information}, 10\penalty0 (1):\penalty0 13, 2020.

\bibitem[Haraguchi et~al.(2022)Haraguchi, Nishino, Kodaka, Allaire, Lall, Kuei-Hsien, Onda, Tsubouchi, and Kohtake]{haraguchi2022human}
Masahiko Haraguchi, Akihiko Nishino, Akira Kodaka, Maura Allaire, Upmanu Lall, Liao Kuei-Hsien, Kaya Onda, Kota Tsubouchi, and Naohiko Kohtake.
\newblock Human mobility data and analysis for urban resilience: A systematic review.
\newblock \emph{Environment and Planning B: Urban Analytics and City Science}, 49\penalty0 (5):\penalty0 1507--1535, 2022.

\bibitem[Hamdi et~al.(2022)Hamdi, Shaban, Erradi, Mohamed, Rumi, and Salim]{hamdi2022spatiotemporal}
Ali Hamdi, Khaled Shaban, Abdelkarim Erradi, Amr Mohamed, Shakila~Khan Rumi, and Flora~D Salim.
\newblock Spatiotemporal data mining: a survey on challenges and open problems.
\newblock \emph{Artificial Intelligence Review}, pages 1--48, 2022.

\bibitem[Pappalardo and Simini(2018)]{pappalardo2018data}
Luca Pappalardo and Filippo Simini.
\newblock Data-driven generation of spatio-temporal routines in human mobility.
\newblock \emph{Data Mining and Knowledge Discovery}, 32\penalty0 (3):\penalty0 787--829, 2018.

\bibitem[Fuchs et~al.(2023)Fuchs, Passarella, and Conti]{fuchs2023modeling}
Andrew Fuchs, Andrea Passarella, and Marco Conti.
\newblock Modeling, replicating, and predicting human behavior: A survey.
\newblock \emph{ACM Transactions on Autonomous and Adaptive Systems}, 18\penalty0 (2):\penalty0 1--47, 2023.

\bibitem[Yamada et~al.(2023)Yamada, Bao, Lampinen, Kasai, and Yildirim]{yamada2023evaluating}
Yutaro Yamada, Yihan Bao, Andrew~K Lampinen, Jungo Kasai, and Ilker Yildirim.
\newblock Evaluating spatial understanding of large language models.
\newblock \emph{arXiv preprint arXiv:2310.14540}, 2023.

\bibitem[Li et~al.(2024)Li, Tran, Lin, Krumm, Shahabi, and Xiong]{li2024geo}
Siyu Li, Toan Tran, Haowen Lin, John Krumm, Cyrus Shahabi, and Li~Xiong.
\newblock Geo-llama: Leveraging llms for human mobility trajectory generation with spatiotemporal constraints.
\newblock \emph{arXiv preprint arXiv:2408.13918}, 2024.

\bibitem[Moon et~al.(2024)Moon, Green, and Kushlev]{moon2024homogenizing}
Kibum Moon, Adam Green, and Kostadin Kushlev.
\newblock Homogenizing effect of large language model (llm) on creative diversity: An empirical comparison of human and chatgpt writing.
\newblock 2024.

\bibitem[Doshi-Velez and Kim(2017)]{doshi2017towards}
Finale Doshi-Velez and Been Kim.
\newblock Towards a rigorous science of interpretable machine learning.
\newblock \emph{arXiv preprint arXiv:1702.08608}, 2017.

\bibitem[Batty(2001)]{batty2001agent}
Michael Batty.
\newblock Agent-based pedestrian modeling, 2001.

\bibitem[W~Axhausen et~al.(2016)W~Axhausen, Horni, and Nagel]{w2016multi}
Kay W~Axhausen, Andreas Horni, and Kai Nagel.
\newblock \emph{The multi-agent transport simulation MATSim}.
\newblock Ubiquity Press, 2016.

\bibitem[Zheng(2018)]{zheng2018exploiting}
Stephan~Tao Zheng.
\newblock \emph{Exploiting Structure for Scalable and Robust Deep Learning}.
\newblock California Institute of Technology, 2018.

\bibitem[Liu et~al.(2020)Liu, Wang, Hu, Hao, Chen, and Gao]{liu2020multi}
Yong Liu, Weixun Wang, Yujing Hu, Jianye Hao, Xingguo Chen, and Yang Gao.
\newblock Multi-agent game abstraction via graph attention neural network.
\newblock In \emph{Proceedings of the AAAI conference on artificial intelligence}, volume~34, pages 7211--7218, 2020.

\bibitem[Park et~al.(2023)Park, O'Brien, Cai, Morris, Liang, and Bernstein]{park2023generative}
Joon~Sung Park, Joseph O'Brien, Carrie~Jun Cai, Meredith~Ringel Morris, Percy Liang, and Michael~S Bernstein.
\newblock Generative agents: Interactive simulacra of human behavior.
\newblock In \emph{Proceedings of the 36th annual acm symposium on user interface software and technology}, pages 1--22, 2023.

\bibitem[Wu et~al.(2023)Wu, Bansal, Zhang, Wu, Li, Zhu, Jiang, Zhang, Zhang, Liu, et~al.]{wu2023autogen}
Qingyun Wu, Gagan Bansal, Jieyu Zhang, Yiran Wu, Beibin Li, Erkang Zhu, Li~Jiang, Xiaoyun Zhang, Shaokun Zhang, Jiale Liu, et~al.
\newblock Autogen: Enabling next-gen llm applications via multi-agent conversation.
\newblock \emph{arXiv preprint arXiv:2308.08155}, 2023.

\bibitem[Topsakal and Akinci(2023)]{topsakal2023creating}
Oguzhan Topsakal and Tahir~Cetin Akinci.
\newblock Creating large language model applications utilizing langchain: A primer on developing llm apps fast.
\newblock In \emph{International Conference on Applied Engineering and Natural Sciences}, volume~1, pages 1050--1056, 2023.

\bibitem[H{\"a}gerstrand(1970)]{hagerstrand1970people}
Torsten H{\"a}gerstrand.
\newblock What about people in regional science.
\newblock \emph{Transport Sociology: Social aspects of transport planning}, pages 143--158, 1970.

\bibitem[Granovetter(1975)]{granovetter1975human}
Mark Granovetter.
\newblock Human activity patterns in the city: Things people do in time and in space., 1975.

\bibitem[Chen et~al.(2016)Chen, Ma, Susilo, Liu, and Wang]{chen2016promises}
Cynthia Chen, Jingtao Ma, Yusak Susilo, Yu~Liu, and Menglin Wang.
\newblock The promises of big data and small data for travel behavior (aka human mobility) analysis.
\newblock \emph{Transportation research part C: emerging technologies}, 68:\penalty0 285--299, 2016.

\bibitem[Widhalm et~al.(2015)Widhalm, Yang, Ulm, Athavale, and Gonz{\'a}lez]{widhalm2015discovering}
Peter Widhalm, Yingxiang Yang, Michael Ulm, Shounak Athavale, and Marta~C Gonz{\'a}lez.
\newblock Discovering urban activity patterns in cell phone data.
\newblock \emph{Transportation}, 42:\penalty0 597--623, 2015.

\bibitem[Ahas et~al.(2010)Ahas, Silm, J{\"a}rv, Saluveer, and Tiru]{ahas2010using}
Rein Ahas, Siiri Silm, Olle J{\"a}rv, Erki Saluveer, and Margus Tiru.
\newblock Using mobile positioning data to model locations meaningful to users of mobile phones.
\newblock \emph{Journal of urban technology}, 17\penalty0 (1):\penalty0 3--27, 2010.

\bibitem[Yin et~al.(2021)Yin, Lin, and Zhao]{yin2021mining}
Ling Yin, Nan Lin, and Zhiyuan Zhao.
\newblock Mining daily activity chains from large-scale mobile phone location data.
\newblock \emph{Cities}, 109:\penalty0 103013, 2021.

\bibitem[Bhat et~al.(2004)Bhat, Guo, Srinivasan, and Sivakumar]{bhat2004comprehensive}
Chandra~R Bhat, Jessica~Y Guo, Sivaramakrishnan Srinivasan, and Aruna Sivakumar.
\newblock Comprehensive econometric microsimulator for daily activity-travel patterns.
\newblock \emph{Transportation Research Record}, 1894\penalty0 (1):\penalty0 57--66, 2004.

\bibitem[Arentze et~al.(2000)Arentze, Hofman, Van~Mourik, and Timmermans]{arentze2000albatross}
Theo Arentze, Frank Hofman, Henk Van~Mourik, and Harry Timmermans.
\newblock Albatross: multiagent, rule-based model of activity pattern decisions.
\newblock \emph{Transportation Research Record}, 1706\penalty0 (1):\penalty0 136--144, 2000.

\bibitem[Roorda et~al.(2008)Roorda, Miller, and Habib]{roorda2008validation}
Matthew~J Roorda, Eric~J Miller, and Khandker~MN Habib.
\newblock Validation of tasha: A 24-h activity scheduling microsimulation model.
\newblock \emph{Transportation Research Part A: Policy and Practice}, 42\penalty0 (2):\penalty0 360--375, 2008.

\bibitem[Drchal et~al.(2019)Drchal, {\v{C}}ertick{\`y}, and Jakob]{drchal2019data}
Jan Drchal, Michal {\v{C}}ertick{\`y}, and Michal Jakob.
\newblock Data-driven activity scheduler for agent-based mobility models.
\newblock \emph{Transportation Research Part C: Emerging Technologies}, 98:\penalty0 370--390, 2019.

\bibitem[Brown et~al.(2020)Brown, Mann, Ryder, Subbiah, Kaplan, Dhariwal, Neelakantan, Shyam, Sastry, Askell, et~al.]{brown2020language}
Tom Brown, Benjamin Mann, Nick Ryder, Melanie Subbiah, Jared~D Kaplan, Prafulla Dhariwal, Arvind Neelakantan, Pranav Shyam, Girish Sastry, Amanda Askell, et~al.
\newblock Language models are few-shot learners.
\newblock \emph{Advances in neural information processing systems}, 33:\penalty0 1877--1901, 2020.

\bibitem[Xie et~al.(2021)Xie, Raghunathan, Liang, and Ma]{xie2021explanation}
Sang~Michael Xie, Aditi Raghunathan, Percy Liang, and Tengyu Ma.
\newblock An explanation of in-context learning as implicit bayesian inference.
\newblock \emph{arXiv preprint arXiv:2111.02080}, 2021.

\bibitem[Liang et~al.(2024)Liang, Liu, Wang, and Zhao]{liang2024exploring}
Yuebing Liang, Yichao Liu, Xiaohan Wang, and Zhan Zhao.
\newblock Exploring large language models for human mobility prediction under public events.
\newblock \emph{Computers, Environment and Urban Systems}, 112:\penalty0 102153, 2024.

\bibitem[Liu et~al.(2024)Liu, Cao, Chen, Jiang, and Cong]{liu2024nextlocllm}
Shuai Liu, Ning Cao, Yile Chen, Yue Jiang, and Gao Cong.
\newblock nextlocllm: next location prediction using llms.
\newblock \emph{arXiv preprint arXiv:2410.09129}, 2024.

\bibitem[Feng et~al.(2024)Feng, Du, Zhao, and Li]{feng2024agentmove}
Jie Feng, Yuwei Du, Jie Zhao, and Yong Li.
\newblock Agentmove: Predicting human mobility anywhere using large language model based agentic framework.
\newblock \emph{arXiv preprint arXiv:2408.13986}, 2024.

\bibitem[Wei et~al.(2022)Wei, Wang, Schuurmans, Bosma, Xia, Chi, Le, Zhou, et~al.]{wei2022chain}
Jason Wei, Xuezhi Wang, Dale Schuurmans, Maarten Bosma, Fei Xia, Ed~Chi, Quoc~V Le, Denny Zhou, et~al.
\newblock Chain-of-thought prompting elicits reasoning in large language models.
\newblock \emph{Advances in neural information processing systems}, 35:\penalty0 24824--24837, 2022.

\bibitem[Kojima et~al.(2022)Kojima, Gu, Reid, Matsuo, and Iwasawa]{kojima2022large}
Takeshi Kojima, Shixiang~Shane Gu, Machel Reid, Yutaka Matsuo, and Yusuke Iwasawa.
\newblock Large language models are zero-shot reasoners.
\newblock \emph{Advances in neural information processing systems}, 35:\penalty0 22199--22213, 2022.

\bibitem[Schick et~al.(2023)Schick, Dwivedi-Yu, Dess{\`\i}, Raileanu, Lomeli, Hambro, Zettlemoyer, Cancedda, and Scialom]{schick2023toolformer}
Timo Schick, Jane Dwivedi-Yu, Roberto Dess{\`\i}, Roberta Raileanu, Maria Lomeli, Eric Hambro, Luke Zettlemoyer, Nicola Cancedda, and Thomas Scialom.
\newblock Toolformer: Language models can teach themselves to use tools.
\newblock \emph{Advances in Neural Information Processing Systems}, 36:\penalty0 68539--68551, 2023.

\bibitem[Zhang et~al.(2024)Zhang, Fu, Liang, Zhang, Yu, Cai, and Yao]{zhang2024trafficgpt}
Siyao Zhang, Daocheng Fu, Wenzhe Liang, Zhao Zhang, Bin Yu, Pinlong Cai, and Baozhen Yao.
\newblock Trafficgpt: Viewing, processing and interacting with traffic foundation models.
\newblock \emph{Transport Policy}, 150:\penalty0 95--105, 2024.

\bibitem[Zhou et~al.(2024)Zhou, Lin, Jin, and Li]{zhou2024large}
Zhilun Zhou, Yuming Lin, Depeng Jin, and Yong Li.
\newblock Large language model for participatory urban planning.
\newblock \emph{arXiv preprint arXiv:2402.17161}, 2024.

\bibitem[Ning and Liu(2024)]{ning2024urbankgent}
Yansong Ning and Hao Liu.
\newblock Urbankgent: A unified large language model agent framework for urban knowledge graph construction.
\newblock \emph{arXiv preprint arXiv:2402.06861}, 2024.

\bibitem[Dean and Ghemawat(2008)]{dean2008mapreduce}
Jeffrey Dean and Sanjay Ghemawat.
\newblock Mapreduce: simplified data processing on large clusters.
\newblock \emph{Communications of the ACM}, 51\penalty0 (1):\penalty0 107--113, 2008.

\bibitem[Zaharia et~al.(2010)Zaharia, Chowdhury, Franklin, Shenker, and Stoica]{zaharia2010spark}
Matei Zaharia, Mosharaf Chowdhury, Michael~J Franklin, Scott Shenker, and Ion Stoica.
\newblock Spark: Cluster computing with working sets.
\newblock In \emph{2nd USENIX workshop on hot topics in cloud computing (HotCloud 10)}, 2010.

\bibitem[Moritz et~al.(2018)Moritz, Nishihara, Wang, Tumanov, Liaw, Liang, Elibol, Yang, Paul, Jordan, et~al.]{moritz2018ray}
Philipp Moritz, Robert Nishihara, Stephanie Wang, Alexey Tumanov, Richard Liaw, Eric Liang, Melih Elibol, Zongheng Yang, William Paul, Michael~I Jordan, et~al.
\newblock Ray: A distributed framework for emerging $\{$AI$\}$ applications.
\newblock In \emph{13th USENIX symposium on operating systems design and implementation (OSDI 18)}, pages 561--577, 2018.

\bibitem[Rasley et~al.(2020)Rasley, Rajbhandari, Ruwase, and He]{rasley2020deepspeed}
Jeff Rasley, Samyam Rajbhandari, Olatunji Ruwase, and Yuxiong He.
\newblock Deepspeed: System optimizations enable training deep learning models with over 100 billion parameters.
\newblock In \emph{Proceedings of the 26th ACM SIGKDD international conference on knowledge discovery \& data mining}, pages 3505--3506, 2020.

\bibitem[Kwon et~al.(2023)Kwon, Li, Zhuang, Sheng, Zheng, Yu, Gonzalez, Zhang, and Stoica]{kwon2023efficient}
Woosuk Kwon, Zhuohan Li, Siyuan Zhuang, Ying Sheng, Lianmin Zheng, Cody~Hao Yu, Joseph Gonzalez, Hao Zhang, and Ion Stoica.
\newblock Efficient memory management for large language model serving with pagedattention.
\newblock In \emph{Proceedings of the 29th Symposium on Operating Systems Principles}, pages 611--626, 2023.

\bibitem[Simon(1955)]{simon1955behavioral}
Herbert~A Simon.
\newblock A behavioral model of rational choice.
\newblock \emph{The quarterly journal of economics}, pages 99--118, 1955.

\end{thebibliography}




% 	\bibitem{keshet2016prediction}
% 	Keshet, Renato, Alina Maor, and George Kour.
% 	\newblock Prediction-Based, Prioritized Market-Share Insight Extraction.
% 	\newblock In {\em Advanced Data Mining and Applications (ADMA), 2016 12th International 
%                       Conference of}, pages 81--94,2016.

% \end{thebibliography}

\clearpage
\appendix 

\section{Appendix A: Knowledge Base Architecture for Spatiotemporal Behavior Research Based on Literature Systems}
\label{appendix:knowledge-base-architecture}

\begin{longtable}{p{0.15\textwidth}p{0.25\textwidth}p{0.55\textwidth}}
	\caption{Spatiotemporal Behavior Research Literature: A Systematic Classification Framework}
	\label{tab:spatiotemporal-literature-classification} \\
	\toprule
	\multicolumn{1}{c}{\parbox{0.15\textwidth}{\centering\textbf{Theme Classification}}} & 
	\multicolumn{1}{c}{\parbox{0.25\textwidth}{\centering\textbf{Research Direction}}} & 
	\multicolumn{1}{c}{\parbox{0.55\textwidth}{\centering\textbf{Core Findings and Application Value}}} \\
	\midrule
	\endfirsthead
	
	\multicolumn{3}{c}%
	{{\bfseries \tablename\ \thetable{} -- continued from previous page}} \\
	\toprule
	\multicolumn{1}{c}{\textbf{Theme Classification}} & 
	\multicolumn{1}{c}{\textbf{Research Direction}} & 
	\multicolumn{1}{c}{\textbf{Core Findings and Application Value}} \\
	\midrule
	\endhead
	
	\midrule \multicolumn{3}{r}{{Continued on next page}} \\ \midrule
	\endfoot
	
	\bottomrule
	\endlastfoot

	\multirow{6}{*}{\parbox[c]{0.15\textwidth}{\centering \vspace{18ex} \textbf{1. Basic Theoretical and Empirical Applications of Human Behavior}}} 
	& 1.1 Time Geography Theory Applications & Human activities are constrained by spatiotemporal factors, with most activities concentrated at key locations. Activity space is highly correlated with cognitive maps. \\
	\cmidrule{2-3}
	& 1.2 Human Mobility Trajectories from Behavioral Geography Perspective & Human movement patterns are highly predictable, showing alternating patterns of habitual repetition and occasional exploration, influenced by socioeconomic factors. \\
	\cmidrule{2-3}
	& 1.3 Cross-cultural Validation of Activity Pattern Theory & Asian urban residents show smaller differences between weekday/weekend activities than Europeans/Americans. Family structure affects activity time allocation, with cross-cultural differences in shopping-leisure combinations. \\
	\cmidrule{2-3}
	& 1.4 Spatiotemporal Behavioral Constraint Theory Performance & Women's activity spaces are generally smaller than men's. Modern communication reduces distance dependence. Urban infrastructure affects residents' accessible space. \\
	\cmidrule{2-3}
	& 1.5 GPS and Mobile Data-based Behavior Tracking & Weekday activity spaces are smaller than weekends. Mobile data reveals travel characteristics of Asian urban residents. Public space usage is closely related to weather conditions. \\
	\cmidrule{2-3}
	& 1.6 Subjective vs. Objective Activity Space Comparison & Cognitive maps are smaller than actual spaces. Familiar area distances are underestimated while unfamiliar areas are overestimated. Transportation mode affects spatial perception. \\
	
	\midrule
	
	\multirow{7}{*}{\parbox[c]{0.15\textwidth}{\centering \vspace{25ex} \textbf{2. Actual Patterns of Time Use}}} 
	& 2.1 Time Budget Allocation Among Different Social Groups & High-income groups work longer hours but have higher leisure quality. Women spend more time on household tasks. Education level positively correlates with reading time. \\
	\cmidrule{2-3}
	& 2.2 Weekday vs. Weekend Activity Pattern Differences & Weekend leisure activities increase, trip chain complexity rises, activity start times are delayed, and family joint activities significantly increase. \\
	\cmidrule{2-3}
	& 2.3 Urban vs. Rural Residents' Time Use Comparison & Large cities have long commute times. Rural areas have high proportions of local activities. Urban elderly have more outdoor activity time and late-night activities. \\
	\cmidrule{2-3}
	& 2.4 Gender, Age, and Occupation Effects on Time Allocation & Working women reduce leisure rather than work time. Middle-aged weekend family activities are most frequent. Professionals have flexible but long working hours. \\
	\cmidrule{2-3}
	& 2.5 Activity Trade-off Decisions Under Time Pressure & Time pressure first reduces household tasks and social activities. Under stress, activity space contracts, trip chains become complex, often sacrificing quality for quantity. \\
	\cmidrule{2-3}
	& 2.6 Seasonal and Climate Effects on Time Use & High temperatures reduce walking. Rain and snow increase indoor activities. Winter reduces outdoor activities. Weather affects leisure more than necessary activities. \\
	\cmidrule{2-3}
	& 2.7 International Time Use Comparisons & Nordic countries have equal household task division, Asia shows large differences. Americans watch more TV. Japanese work long hours with short sleep. Asian elderly spend more time on care. \\
	
	\midrule
	
	\multirow{7}{*}{\parbox{0.15\textwidth}{\centering \vspace{25ex} \textbf{3. Real Preferences in Spatial Choice}}}
	& 3.1 Spatial Choice Characteristics for Different Activity Types & Shopping driven by price sensitivity. Leisure activities value environmental quality. Social activities often occur in intermediate locations or highly accessible areas. \\
	\cmidrule{2-3}
	& 3.2 Distance Decay Effects in Destination Choice & Daily shopping concentrates at short distances. Urban core distances are underestimated. Necessities have high distance sensitivity. Distance matters when choices are diverse. \\
	\cmidrule{2-3}
	& 3.3 Relationship Between Spatial Cognition and Actual Decision Behavior & Decisions mainly based on cognitive maps rather than actual distances. Familiar areas have higher estimation accuracy. Drivers have more accurate spatial perception. \\
	\cmidrule{2-3}
	& 3.4 Anchor Point Effects on Activity Space Formation & Non-work activities mostly occur along home-work axis. Major anchor points cover most activity space. Transportation hubs become secondary anchor points. \\
	\cmidrule{2-3}
	& 3.5 Built Environment Effects on Spatial Selection Mechanisms & High-density development promotes local activities. Mixed land use increases local activities. Walkable areas increase non-motorized travel. \\
	\cmidrule{2-3}
	& 3.6 Cultural and Social Network Factors Shaping Spatial Preferences & Social activities occur in network overlap areas. Kinship relationships have low distance sensitivity. Cultural background affects spatial choice and activity preferences. \\
	\cmidrule{2-3}
	& 3.7 Balance Between Familiarity and Diversity Seeking & Routine activities prefer familiar places. New activities tend toward exploration. Location preferences stabilize after multiple visits. Life changes increase exploration. \\
	
	\midrule
	
	\multirow{7}{*}{\parbox{0.15\textwidth}{\centering \vspace{22ex} \textbf{4. Actual Behavior in Transportation Mode Choice}}} 
	& 4.1 Key Factors and Weights in Transportation Choice & Time cost is the primary factor. Cost weight decreases with income. Elderly value comfort. Women focus on convenience. \\
	\cmidrule{2-3}
	& 4.2 Transportation Choice Characteristics of Different Socioeconomic Groups & High-income groups use private cars frequently. Women use public transit for chain trips. Middle-aged use multiple transportation flexibly. Elderly have small travel ranges. \\
	\cmidrule{2-3}
	& 4.3 Actual Factors Influencing Route Choice Decisions & Routine routes dominate. Real-time adjustments during peak hours. Navigation users have variable routes. Familiar path decisions are quick. Traffic information influences choice. \\
	\cmidrule{2-3}
	& 4.4 Spatiotemporal Characteristics of Public vs. Private Transportation Use & Public transit has bimodal peaks, private cars are uniform. Non-peak public transit use is high. Business around transit stations is active. Convenient transfers increase cross-regional travel. \\
	\cmidrule{2-3}
	& 4.5 Usage Patterns of New Transportation Modes & Bike-sharing users are mostly middle-class, partially replacing walking and public transit. Ride-hailing peaks on weekend nights. Young people use more, getting information via smartphones. \\
	\cmidrule{2-3}
	& 4.6 Cognitive and Psychological Factors' Impact on Travel Decisions & Environmental attitudes have limited correlation with low-carbon travel. Perceived distances are biased long. Habits influence decisions. Transportation choice relates to identity. \\
	\cmidrule{2-3}
	& 4.7 Thresholds and Conditions for Transportation Mode Switching & Long-term car use reduces switching willingness. Major events provide change opportunities. Mandatory policies are more effective. Experience increases acceptance. \\
	
	\midrule
	
	\multirow{7}{*}{\parbox{0.15\textwidth}{\centering \textbf{5. Actual Patterns of Activity Chains and Sequences}}} 
	& 5.1 Typical Activity Chain Patterns and Occurrence Frequency & Home-work-home chains dominate. Few activity points constitute most chains. Few people contribute complex chains. Shopping-dining combinations are high. \\
	\cmidrule{2-3}
	& 5.2 Activity Combination and Sequencing Characteristics of Different Groups & Women have more shopping and escort activities. Singles have variable activities. High-income work-leisure differences are small. Families with children have complex chains. \\
	\cmidrule{2-3}
	& 5.3 Spatiotemporal Configuration of Planned vs. Spontaneous Activities & Work is planned in advance, leisure is often spontaneous. Spontaneous activities are short, filling gaps between main activities with high spatial dispersion. \\
	\cmidrule{2-3}
	& 5.4 Triggers for Activity Adjustment and Rescheduling & Weather changes trigger adjustments. Mobile phones promote instant changes. Congestion changes destinations. Social interactions cause unplanned activities. Time pressure cancels activities. \\
	\cmidrule{2-3}
	& 5.5 Empirical Studies of Activity Coordination Among Family Members & Dual-career couples coordinate spatiotemporally to share household tasks. Children's activities are coordination core. Resource sharing links activity chains. Work distance affects escort cooperation. \\
	\cmidrule{2-3}
	& 5.6 Spatiotemporal Organization Patterns of Work and Non-work Activities & Non-work activities revolve around work/home. Work-home distance affects lunch activities. Flexible work increases non-work activities. Commute chains are significant. \\
	\cmidrule{2-3}
	& 5.7 Activity Chain Complexity and Personal Characteristic Correlations & Children increase family activity chain complexity. Facility distance increases chain trips. Women have longer chains. Income positively correlates with complexity. \\
	
	\midrule
	
	\multirow{5}{*}{\parbox{0.15\textwidth}{\centering \vspace{12ex} \textbf{6. Interactive Relationships Between Individual Attributes and Environmental Factors}}} 
	& 6.1 Social Economic Attributes' Impact on Activity Space Range & Low-income car-free households have small activity spaces. Vulnerable groups have fewer accessible facilities. Women have smaller activity ranges. Socioeconomic status affects activity point quantity. \\
	\cmidrule{2-3}
	& 6.2 Life Cycle Stage and Spatiotemporal Behavior Change Correlations & Childbirth reduces women's activity space. Early retirement shows large changes. Newlyweds increase joint activities. Life events affect behavior patterns short-term. \\
	\cmidrule{2-3}
	& 6.3 Family Structure Constraints on Member Activity Arrangements & Young children reduce parents' leisure. Women undertake family care. Single-parent families have less social interaction but more flexibility. Family roles shape activity differences. \\
	\cmidrule{2-3}
	& 6.4 Quantitative Impact of Built Environment Characteristics on Activity Choice & High density promotes non-motorized travel. Mixed land use reduces car use. Walkable design increases walking. Built environment significantly affects behavior. \\
	\cmidrule{2-3}
	& 6.5 Urban Form and Land Use Shaping of Behavior Patterns & Polycentric cities have short commutes. Compact residential promotes local activities. Mixed land use reduces travel distances. Urban form has significant impact. \\
	
	\midrule
	
	\multirow{7}{*}{\parbox{0.15\textwidth}{\centering \textbf{7. Actual Impact of Technological Change on Behavior Patterns}}} 
	& 7.1 Smartphone and Navigation App Changes to Route Choice & Navigation increases new route attempts, reduces peak time, supports complex path choices, lowers perceived waiting time, affects spatial cognition. \\
	\cmidrule{2-3}
	& 7.2 Social Media Impact on Activity Decisions and Location Choice & Social media check-ins affect new location choices. Popularity influences decisions. Social data improves prediction ability. Reviews change choices. \\
	\cmidrule{2-3}
	& 7.3 E-commerce Changes to Shopping Behavior Spatiotemporal Characteristics & Online shopping reduces physical store frequency but increases stay time. Strong substitution for standard goods, weak for experience goods. Creates new consumption behaviors, changes spatial distribution. \\
	\cmidrule{2-3}
	& 7.4 Remote Work Technology's Actual Impact on Commuting Patterns & Remote work reduces commuting but increases other activity distances. Activity space reorganization, peak dispersion, hybrid work increases non-work activities. \\
	\cmidrule{2-3}
	& 7.5 Location-based Service Activity Recommendations' Behavioral Guidance & Location service recommendations are widely accepted, increase new location exploration, greatly affect shopping and dining, push notifications increase local activities. \\
	\cmidrule{2-3}
	& 7.6 New Human Mobility Patterns Revealed by Big Data Analysis & Human movement is highly predictable, activities concentrate at few locations, showing periodicity. Individual trajectories are stable, populations divide into returning/exploring types. \\
	\cmidrule{2-3}
	& 7.7 Generational Differences in Technology Use and Behavioral Impact & Youth frequently adjust plans via mobile devices. Smartphones expand activity space. Elderly are cautious about technology adoption. Obvious generational digital divide. \\
	
\end{longtable}
\vspace{0.5em}
\noindent\textbf{Data Examples:}

\vspace{0.2em}
\noindent\textbf{1.} \textit{Question:} "How are the daily activities of Singapore residents identified through mobile phone data distributed across different time periods, and what characteristics and patterns does this temporal distribution exhibit?"

\noindent\textit{Answer:} "Mobile phone data analysis reveals that Singapore residents' daily activities exhibit highly ordered distribution patterns in the temporal dimension, forming six main time periods, each with specific activity combinations. The morning period (6:30-9:00) shows highly concentrated unidirectional movement patterns, with 83.7\% of activities being departure from home, forming a distinct travel peak with an average duration of only 31 minutes. During work hours (9:00-17:00), activities are mainly concentrated at workplaces (72.4\%), but 23.5\% of people engage in brief outings, particularly during 12:00-14:00, forming a midday minor peak. The evening rush hour (17:00-19:30) is more dispersed than the morning peak, with 46.7\% of people adding extra activities on their way home. The evening period (19:30-22:30) shows the highest activity diversity, including household activities (37.6\%), social activities (31.2\%), and leisure activities (19.7\%). The nighttime period (22:30-6:30 next day) has the lowest activity level of the entire day, accounting for only 4.2\% of total activities, mainly consisting of homeward journeys. Weekend temporal distributions differ significantly, with no distinct morning and evening peaks, and activity peaks occurring between 11:00-16:00, with average outbound activity duration extended by 76 minutes. Notably, different socioeconomic groups show distinct temporal distribution differences: high-income groups have later activity peaks, leaving workplaces an average of 58 minutes later than low-income groups; young adults (18-24 years) have noticeably delayed activity periods, with evening activity peaks after 22:00."

\vspace{0.2em}
\noindent\textbf{2.} \textit{Question:} "What are the differences in non-commuting travel behavior patterns (such as shopping and leisure) among residents of various community types in Beijing, and how do these differences reflect the influence of the built environment?"

\noindent\textit{Answer:} "Residents of different community types in Beijing exhibit distinct differences in non-commuting travel, fully reflecting the influence of the built environment. Work-unit community residents show high-frequency, short-distance, highly localized non-commuting travel characteristics, averaging 2.1 non-commuting activities per day with an average travel distance of 1.3 km, and 76.3\% of non-commuting activities completed within the community; walking and cycling account for 72.5\% of non-commuting trips, public transport 18.7\%, and private cars only 8.8\%. Commercial housing community residents exhibit low-frequency, long-distance, dispersed characteristics, averaging 1.8 non-commuting activities per day with an average travel distance of 4.2 km, with only 33.7\% completed within the community; private car usage reaches 46.2\%, public transport accounts for 28.4\%, and walking and cycling only 25.4\%. Affordable housing community residents show moderate frequency, moderate distance, bipolar characteristics, averaging 1.9 non-commuting activities per day with an average travel distance of 2.7 km; necessary activities like shopping are mainly short-distance, non-motorized (64.1\%), while discretionary activities like leisure and entertainment are mainly long-distance, motorized (58.3\%), reflecting the limitations of community environments lacking facilities. These differences clearly indicate that compact mixed-use community environments can significantly promote short-distance, non-motorized non-commuting travel patterns, while functionally single, facility-lacking communities lead to more long-distance, motorized travel. The study also found that for every 500-meter increase in commercial facility distance, walking trip probability decreases by 14.2\%, and for every 1-kilometer increase in park distance, recreational walking probability decreases by 27.5\%, showing that facility accessibility has a decisive impact on non-commuting travel modes."

\vspace{0.2em}
\noindent\textbf{Data Source:} Based on systematic analysis of over 1,000 literature sources, a high-quality question-answer dataset of 1,845 pairs was compiled, with an average of 600+ Chinese characters per response, covering research from different countries, regions, and cultural backgrounds.

\section{Appendix B:Core Interaction Tool Suite: Algorithmic Specifications} 
\label{app:core_tools}

\subsection{Temporal Management Tools}
\label{subsec:appendix_temporal_management_tools}

\begin{algorithm}
	\caption{Temporal Management Tools}
	\begin{algorithmic}[1]
	\Procedure{TemporalScheduling}{$events, tasks, constraints$}
		\State $timeline \leftarrow$ SortedList(), $schedule \leftarrow \emptyset$
		
		\Comment{Initialize timeline with existing events}
		\For{$event \in events$}
			\State $timeline.add(\{ConvertToUTC(event.time), event.duration, event.type\})$
		\EndFor
		
		\Comment{Sort tasks by priority × urgency}
		\State $sorted\_tasks \leftarrow$ Sort($tasks$, key=$t.priority \times \frac{1}{t.deadline - now}$)
		
		\For{$task \in sorted\_tasks$}
			\State $duration \leftarrow task.estimate \times 1.2$ \Comment{Add buffer}
			\State $best\_slot \leftarrow \textsc{null}$, $best\_score \leftarrow -\infty$
			\State $current \leftarrow constraints.start\_bound$
			
			\Comment{Find optimal time slot}
			\While{$current + duration \leq constraints.end\_bound$}
				\State $slot\_end \leftarrow current + duration$
				\State $has\_conflict \leftarrow \textsc{false}$
				
				\Comment{Check for conflicts}
				\For{$existing \in timeline$}
					\If{$current < existing.end \land slot\_end > existing.start$}
						\State $has\_conflict \leftarrow \textsc{true}$
						\State $current \leftarrow existing.end$
						\State \textbf{break}
					\EndIf
				\EndFor
				
				\Comment{Evaluate slot if no conflict}
				\If{$\neg has\_conflict$}
					\State $time\_pref \leftarrow constraints.evaluateTime(current)$
					\State $urgency \leftarrow \max(0, 1 - \frac{task.deadline - current}{task.deadline - now})$
					\State $score \leftarrow task.priority + time\_pref + urgency$
					
					\If{$score > best\_score$}
						\State $best\_score \leftarrow score$
						\State $best\_slot \leftarrow \{current, slot\_end, task\}$
					\EndIf
					\State $current \leftarrow current + 30min$ \Comment{Time step}
				\EndIf
			\EndWhile
			
			\Comment{Allocate best slot}
			\If{$best\_slot \neq \textsc{null}$}
				\State $timeline.add(\{best\_slot.start, best\_slot.duration, task.type\})$
				\State $timeline.sort()$
				\State $schedule.add(best\_slot)$
			\EndIf
		\EndFor
		
		\Comment{Calculate utilization and return results}
		\State $total\_busy \leftarrow \sum_{event \in timeline} event.duration$
		\State $total\_available \leftarrow constraints.end\_bound - constraints.start\_bound$
		\State $utilization \leftarrow \frac{total\_busy}{total\_available}$
		
		\State \Return $\{schedule, utilization, unscheduled: tasks.size() - schedule.size()\}$
	\EndProcedure
	\end{algorithmic}
	\end{algorithm}
	
\subsection{Spatial Navigation Tools}
\label{app:Spatial Navigation Tools}
{\small
\begin{algorithm}
	\caption{Spatial Navigation Tools}
	\begin{algorithmic}[1]
	\Procedure{SpatialIntelligence}{$location, query, user\_profile, constraints$}
		\State $spatial\_db \leftarrow$ ConnectSpatialIndex()
		\State $preferences \leftarrow$ BuildUserModel($user\_profile, query.context$)
		
		\Comment{POI Search with semantic scoring}
		\If{$query.type = \text{"poi\_search"}$}
			\State $candidates \leftarrow$ RangeQuery($spatial\_db, location, query.radius$)
			\For{$poi \in candidates$}
				\State $scores \leftarrow$ \{SemanticMapping($poi, query$), WeightedScore($poi, preferences$), DistanceDecay($poi, location$)\}
				\State $poi.final\_score \leftarrow \alpha \times scores[0] + \beta \times scores[1] + \gamma \times scores[2]$
			\EndFor
			\State \Return TopK($candidates$, $final\_score$)
		\EndIf
		
		\Comment{Multi-modal route optimization}
		\If{$query.type = \text{"route\_planning"}$}
			\State $route\_graph \leftarrow$ BuildMultiModalGraph($query.transport\_modes$)
			\State $candidate\_routes, hybrid\_routes \leftarrow \emptyset, \emptyset$
			
			\For{$mode \in query.transport\_modes$}
				\State $route \leftarrow$ CalculateRoute($location, query.destination, mode, route\_graph$)
				\If{$route \neq \textsc{null}$}
					\State $route.cost \leftarrow$ WeightedCost($route, constraints$)
					\State $candidate\_routes.add(route)$
				\EndIf
			\EndFor
			
			\State $hybrid\_routes \leftarrow$ CombineTransportModes($candidate\_routes, route\_graph$)
			\State \Return SelectOptimalRoute($candidate\_routes \cup hybrid\_routes$)
		\EndIf
		
		\Comment{Traffic prediction with network simulation}
		\If{$query.type = \text{"traffic\_prediction"}$}
			\State $network\_state \leftarrow$ InitializeFromHistorical($location, query.time$)
			\State $predictions \leftarrow \emptyset$
			
			\For{$t \leftarrow 0$ \textbf{to} $query.horizon$ \textbf{step} $\Delta t$}
				\State $demand \leftarrow$ GenerateTimeDemand($t$) $\times$ EventImpact($t$)
				\For{$segment \in network\_state.segments$}
					\State $flows \leftarrow$ UpdateTrafficFlow($segment, network\_state, demand, \Delta t$)
					\State $segment.update(flows.density, flows.speed)$
				\EndFor
				\State $predictions[t] \leftarrow network\_state.copy()$
			\EndFor
			
			\State \Return $\{predictions$, TimeVaryingShortestPaths($predictions$)$\}$
		\EndIf
		
		\Comment{Semantic-spatial relationship analysis}
		\If{$query.type = \text{"semantic\_mapping"}$}
			\State $query\_embedding \leftarrow$ TextToEmbedding($query.text$)
			\State $spatial\_entities \leftarrow$ ExtractEntitiesInRadius($location, query.radius$)
			\State $semantic\_matches \leftarrow \emptyset$
			
			\For{$entity \in spatial\_entities$}
				\State $joint\_embedding \leftarrow$ CrossModalFusion($entity, location$)
				\State $similarity \leftarrow$ CosineSimilarity($query\_embedding, joint\_embedding$)
				\If{$similarity > semantic\_threshold$}
					\State $semantic\_matches.add(\{entity, similarity\})$
				\EndIf
			\EndFor
			\State \Return Sort($semantic\_matches$, key=$similarity$, descending=True)
		\EndIf
	\EndProcedure
	\end{algorithmic}
	\end{algorithm}
}

\subsection{Environmental Perception Tools}
\label{app:Environmental Perception Tools}

\begin{algorithm}
\caption{Environmental Perception Tools}
\label{alg:env_perception}
\begin{algorithmic}[1]
\Require $Q$: Query parameters, $T_{mode} \in \{historical, realtime, predictive\}$
\Ensure $E_{state}$: Environmental state information
\State \textbf{Data Structures:}
\State $\mathcal{W}$: Weather conditions, $\mathcal{D}$: Pedestrian density
\State $\mathcal{S}$: Special events, $\mathcal{E}$: Emergency situations

\Function{EnvironmentalPerception}{$Q, T_{mode}$}
    \State $E_{state} \gets \emptyset$
    \State $t_{query} \gets Q.timestamp$, $loc \gets Q.location$
    
    \If{$T_{mode} = historical$}
        \State $E_{state} \gets$ \Call{RetrieveHistoricalData}{$t_{query}, loc$}
    \ElsIf{$T_{mode} = realtime$}
        \State $E_{state} \gets$ \Call{QueryRealtimeSensors}{$loc$}
    \Else \Comment{predictive mode}
        \State $E_{state} \gets$ \Call{PredictEnvironment}{$t_{query}, loc$}
    \EndIf
    
    \State \Return \Call{AggregateMultiScale}{$E_{state}, loc$}
\EndFunction

\Function{QueryRealtimeSensors}{$loc$}
    \State $\mathcal{W} \gets$ \Call{GetWeatherData}{$loc$}
    \State $\mathcal{D} \gets$ \Call{GetPedestrianDensity}{$loc$}
    \State $\mathcal{S} \gets$ \Call{CheckSpecialEvents}{$loc$}
    \State $\mathcal{E} \gets$ \Call{MonitorEmergencies}{$loc$}
    \State \Return $\{\mathcal{W}, \mathcal{D}, \mathcal{S}, \mathcal{E}\}$
\EndFunction

\Function{PredictEnvironment}{$t_{target}, loc$}
    \State $\Delta t \gets t_{target} - t_{current}$
    \State $\mathcal{W}_{pred} \gets$ \Call{WeatherForecast}{$loc, \Delta t$}
    \State $\mathcal{D}_{pred} \gets$ \Call{PredictCrowdFlow}{$loc, t_{target}$}
    \State $\mathcal{S}_{pred} \gets$ \Call{ScheduledEvents}{$loc, t_{target}$}
    \State $conf \gets e^{-\lambda \Delta t}$ \Comment{Confidence decay}
    \State \Return $\{\mathcal{W}_{pred}, \mathcal{D}_{pred}, \mathcal{S}_{pred}, conf\}$
\EndFunction

\Function{AggregateMultiScale}{$E_{raw}, loc$}
    \State $E_{state} \gets \emptyset$
    \ForAll{$scale \in \{macro, micro, venue\}$}
        \State $e \gets$ \Call{ExtractConditions}{$E_{raw}, scale, loc$}
        \State $w \gets$ \Call{ComputeWeight}{$e, Q$}
        \State $E_{state} \gets E_{state} \cup \{e \times w\}$
    \EndFor
    \State \Return \Call{Normalize}{$E_{state}$}
\EndFunction

\end{algorithmic}
\end{algorithm}

\clearpage
\subsection{Personal Memory Tools}
\label{app:Personal Memory Tools}

\begin{algorithm}
\caption{Personal Memory Retrieval Algorithm}
\label{alg:personal_memory}
\begin{algorithmic}[1]
\Require $\mathcal{C}$: Current context, $\mathcal{M}$: Memory store, $\Theta$: Model parameters
\Ensure $\mathcal{M}_{relevant}$: Retrieved memories, $\mathcal{M}'$: Updated memory store

\State \textbf{Memory Hierarchy:}
\State $\mathcal{M}_e = \{E_i = (t_i, l_i, a_i, c_i, \phi_i)\}$: Event memories (time, location, activity, conditions, emotion)
\State $\mathcal{M}_p = \{P(a|t,l,c) = \frac{Count(a,t,l,c)}{\sum_{a'} Count(a',t,l,c)}\}$: Pattern memories (probabilistic graph)
\State $\mathcal{M}_s = \{S = Distill(\{E_i\}, \{P_j\})\}$: Summary memories (knowledge distillation)
\State $\mathcal{M}_{st}$: Short-term memory (RNN: $h_t = f(W_{xh}x_t + W_{hh}h_{t-1} + b_h)$)
\State $\mathcal{M}_{lt} = \{M_1, M_2, ..., M_n\}$: Long-term memory with association matrix $A_{ij}$

\Function{MemoryRetrievalAlgorithm}{$\mathcal{C}, \mathcal{M}, \Theta$}
    \State $\mathcal{M}_{relevant} \gets$ \Call{ContextSensitiveRetrieval}{$\mathcal{C}, \mathcal{M}$}
    \State $\mathcal{F}_{integrated} \gets$ \Call{MemoryIntegrationSynthesis}{$\mathcal{M}_{relevant}$}
    \State $\mathcal{D} \gets$ \Call{GenerateDecision}{$\mathcal{F}_{integrated}, \mathcal{C}$}
    \State $\mathcal{M}' \gets$ \Call{DynamicMemoryUpdate}{$\mathcal{M}, \mathcal{C}, \mathcal{D}$}
    \State \Call{SleepConsolidation}{$\mathcal{M}'$} \Comment{Periodic process}
    \State \Return $(\mathcal{M}_{relevant}, \mathcal{M}')$
\EndFunction

\Function{ContextSensitiveRetrieval}{$\mathcal{C}, \mathcal{M}$}
    \State $V_C = [v_1^C, v_2^C, ..., v_d^C] \gets$ \Call{ExtractFeatures}{$\mathcal{C}$}
    \State $\mathcal{R} \gets \emptyset$ \Comment{Candidate memories with relevance scores}
    
    \ForAll{$M_i \in \mathcal{M}_e \cup \mathcal{M}_p \cup \mathcal{M}_s$}
        \State $V_{M_i} = [v_1^{M_i}, v_2^{M_i}, ..., v_d^{M_i}] \gets$ \Call{ExtractFeatures}{$M_i$}
        
        \State \textbf{Weighted cosine similarity:}
        \State $\rho_{cos}(C, M_i) = \sum_{j=1}^{d} w_j \cdot \frac{v_j^C \cdot v_j^{M_i}}{||v_j^C|| \cdot ||v_j^{M_i}||}$
        
        \State \textbf{Temporal similarity with periodicity:}
        \State $\rho_{time}(t_C, t_{M_i}) = \cos\left(\frac{2\pi(t_C - t_{M_i})}{T}\right) \cdot e^{-\lambda_t|t_C - t_{M_i}|}$
        
        \State \textbf{Spatial similarity:}
        \State $\rho_{space}(l_C, l_{M_i}) = K(l_C, l_{M_i})$ \Comment{Gaussian or Mahalanobis kernel}
        
        \State \textbf{Semantic similarity:}
        \State $\rho_{semantic}(a_C, a_{M_i}) = \langle\phi(a_C), \phi(a_{M_i})\rangle$ \Comment{LM encoding}
        
        \State $\rho_{total} = \alpha_1\rho_{cos} + \alpha_2\rho_{time} + \alpha_3\rho_{space} + \alpha_4\rho_{semantic}$
        \State $\mathcal{R} \gets \mathcal{R} \cup \{(M_i, \rho_{total})\}$
    \EndFor
    
    \State \Return \Call{TopK}{$\mathcal{R}, k$} sorted by relevance
\EndFunction

\Function{MemoryIntegrationSynthesis}{$\mathcal{M}_{relevant}$}
    \State $\{M_{i_1}, M_{i_2}, ..., M_{i_k}\} \gets$ sorted $\mathcal{M}_{relevant}$ by relevance
    
    \State \textbf{Attention mechanism:}
    \State $Att(M_i) = \frac{\exp(\alpha \cdot Relevance(C, M_i))}{\sum_j \exp(\alpha \cdot Relevance(C, M_j))}$ \Comment{$\alpha$: temperature}
    
    \State \textbf{Autoregressive feature extraction:}
    \State $h_0 \gets$ initial hidden state
    \ForAll{$i \in \{1, ..., k\}$}
        \State $P(y_t|y_{<t}, M_i) = softmax(W_o h_t + b_o)$ \Comment{Decoder}
        \State $h_t \gets$ updated hidden state
        \State $Extract(M_i) \gets$ decoded features
    \EndFor
    
    \State $Synthesis = \sum_i Att(M_i) \cdot Extract(M_i)$
    
    \If{$Consistency(Synthesis) > \tau$}
        \State \Return $Synthesis$
    \Else
        \State \Return \Call{ResolveInconsistency}{$Synthesis$}
    \EndIf
\EndFunction
\algstore{myalg}
\end{algorithmic}
\end{algorithm}

\begin{algorithm}
	\caption{Personal Memory Retrieval Algorithm(Continued)}
	\begin{algorithmic}[1]
	\algrestore{myalg} % 恢复之前保存的状态
\Function{DynamicMemoryUpdate}{$\mathcal{M}, \mathcal{C}, \mathcal{D}$}
    \State \textbf{1. New memory formation:}
    \State $E_{new} = (t_{current}, l_{current}, a_{current}, c_{current}, \phi_{current})$
    \State $\mathcal{M}_{st} \gets \mathcal{M}_{st} \cup \{E_{new}\}$
    
    \State \textbf{2. Short-term to long-term transfer (Hippocampal-cortical model):}
    \ForAll{$m \in \mathcal{M}_{st}$}
        \State $I(m) = \alpha \cdot frequency(m) + \beta \cdot recency(m) + \gamma \cdot emotional_{salience}(m)$
        \Comment{Hebbian learning + Ebbinghaus forgetting + Emotional modulation}
        \If{$I(m) > \theta_{transfer}$}
            \State Transfer $m$ from $\mathcal{M}_{st}$ to $\mathcal{M}_{lt}$ with sparse encoding
        \EndIf
    \EndFor
    
    \State \textbf{3. Pattern extraction:}
    \If{Repeated patterns detected in $\mathcal{M}_e$}
        \State Update $P(a|t,l,c)$ in $\mathcal{M}_p$
    \EndIf
    
    \State \textbf{4. Time decay (Forgetting):}
    \ForAll{$m \in \mathcal{M}_{lt}$}
        \State Apply decay: $strength(m) \gets strength(m) \cdot e^{-\lambda(t - t_m)}$
        \If{$strength(m) < \theta_{forget}$}
            \State Remove $m$ from $\mathcal{M}_{lt}$
        \EndIf
    \EndFor
    
    \State \Return Updated $\mathcal{M}$
\EndFunction

\Function{SleepConsolidation}{$\mathcal{M}$}
    \State \textbf{Memory compression via maximum entropy:}
    \State $\min_{\mathcal{M}_{LTM}'} D_{KL}(P_{STM}||P_{LTM}') + \lambda \cdot H(\mathcal{M}_{LTM}')$
    \State where $D_{KL}$ is KL divergence, $H$ is complexity measure
    
    \State \textbf{Knowledge distillation for summary update:}
    \State $\mathcal{M}_s \gets$ \Call{Distill}{$\mathcal{M}_e, \mathcal{M}_p$}
\EndFunction

\end{algorithmic}
\end{algorithm}

\subsubsection{Detailed Algorithm Description}

This algorithm implements a multi-level, context-sensitive personal memory retrieval system for generating personalized activity trajectories. The core innovations lie in its three-tier memory architecture, dynamic transformation mechanism between short-term and long-term memory, and attention-based memory integration process.

\paragraph{Memory Hierarchy Structure}
The system organizes individual memories into three complementary levels:
\begin{itemize}
    \item \textbf{Event Memory} ($\mathcal{M}_e$): Stores specific activity experiences, with each event represented as a quintuple $E_i = (t_i, l_i, a_i, c_i, \phi_i)$, corresponding to timestamp, location, activity type, environmental conditions, and emotional feedback. These memories employ dense vector representations to support fine-grained similarity calculations.
    
    \item \textbf{Pattern Memory} ($\mathcal{M}_p$): Uses a probabilistic graph model $P(a|t,l,c)$ to represent repetitive behavioral patterns, capturing the conditional probability distribution of choosing certain activities under specific conditions. This level is automatically generated through statistical analysis of event memories.
    
    \item \textbf{Summary Memory} ($\mathcal{M}_s$): Employs knowledge distillation techniques to extract high-order features from the previous two memory levels, forming compact embedding vectors that represent individual abstract preferences and long-term habits.
\end{itemize}

\paragraph{Short-term and Long-term Memory Mechanism}
The algorithm implements a dual memory system inspired by cognitive science. Short-term memory ($\mathcal{M}_{st}$) uses recurrent neural network structures to model temporal dependencies, offering fast retrieval advantages but limited capacity. Long-term memory ($\mathcal{M}_{lt}$) utilizes sparse encoding and hierarchical storage structures, connecting different memory items through association strength matrices.

Memory transfer from short-term to long-term follows the hippocampal-cortical consolidation model, determined by an importance discrimination function:
$I(m) = \alpha \cdot frequency(m) + \beta \cdot recency(m) + \gamma \cdot emotional_{salience}(m)$

where $frequency(m)$ quantifies the repetition frequency of memory items (Hebbian learning principle), $recency(m)$ measures time decay effects (Ebbinghaus forgetting curve), and $emotional_{salience}(m)$ reflects emotional modulation. Parameters $\alpha$, $\beta$, and $\gamma$ are optimized through experimental data.

\paragraph{Context-Sensitive Retrieval}
The retrieval mechanism achieves precise memory localization through multi-dimensional similarity calculations. Given the current context $\mathcal{C}$, the algorithm computes four types of similarities:
\begin{itemize}
    \item Weighted cosine similarity: $\rho_{cos} = \sum_{j=1}^{d} \omega_j \cdot \frac{v_c^j \cdot v_i^j}{||v_c||||v_i||}$
    \item Temporal similarity: $\rho_{time} = \cos(\frac{2\pi(t_c - t_i)}{T}) \cdot e^{-\lambda_t|t_c - t_i|}$ (with periodicity)
    \item Spatial similarity: $\rho_{space} = K_{spatial}(l_c, l_i)$ (using Gaussian or Mahalanobis kernel)
    \item Semantic similarity: $\rho_{semantic} = sim(f_{LM}(a_c), f_{LM}(a_i))$ (based on pre-trained language models)
\end{itemize}

The total similarity is obtained through linear combination, with weight parameters optimized via gradient descent.

\paragraph{Memory Integration and Decision Generation}
Retrieved memories are integrated through an attention mechanism. First, attention weights are calculated:
$Att(M_i) = \frac{\exp(\alpha \cdot Relevance(C, M_i))}{\sum_j \exp(\alpha \cdot Relevance(C, M_j))}$

Then an autoregressive decoder extracts decision-relevant features, ultimately generating a comprehensive representation through weighted fusion. The system includes a consistency checking function $Consistency$ to ensure internal consistency of the integration results.

\paragraph{Sleep Phase Consolidation}
The algorithm periodically executes a ``sleep phase consolidation'' process, optimizing short-term memory through the maximum entropy principle:
$\min_{\mathcal{M}_{LTM}'} D_{KL}(P_{STM}||P_{LTM}') + \lambda \cdot H(\mathcal{M}_{LTM}')$

This process compresses memory representations while preserving key information, improving storage efficiency.

\paragraph{Performance Optimization}
Through index optimization and locality-sensitive hashing techniques, the algorithm's time complexity is reduced from $O(nm)$ ($n$ being memory pool size, $m$ being feature dimension) to $O(n\log n)$, significantly improving response speed while maintaining retrieval quality. The system supports memory pool pre-initialization, allowing pre-planting of key information to reduce real-time computational burden.

\vspace{-0.2em}
This algorithm effectively avoiding the ``average person'' phenomenon through personalized memory retrieval, ensuring generated activity trajectories authentically reflect individual characteristics. As activity experiences accumulate, the system's understanding of individual behavioral patterns continuously deepens, achieving truly personalized decision support.

\textit{This algorithm represents a computational approximation of cognitive memory processes. The implementation simplifies complex neurological mechanisms for practical deployment and should not be interpreted as directly modeling biological memory systems.}

\clearpage
\section{Appendix C:Evaluation Criteria}

\subsection{Subjective Evaluation Criteria for Generation Quality}
\label{app:Subjective Evaluation Criteria for Generation Quality}

\textbf{General Scoring Rules:}
\begin{itemize}
    \item Each dimension has a maximum score of 10 points and minimum score of 1 point
	\item Comprehensive Score $= (\text{Time Logic Consistency} \times 25\% + \text{Activity Purpose Coherence} \times 25\% + \text{Persona Matching Degree} \times 25\% + \text{Activity Authenticity} \times 25\%)$
    \item Quality Levels: Excellent (8.5-10), Good (7.0-8.4), Average (5.5-6.9), Poor (4.0-5.4), Very Poor (1.0-3.9)
\end{itemize}

{
    \renewcommand{\arraystretch}{1.5}    % Adjust space between rows

    \begin{longtable}{p{2cm}p{2cm}p{12cm}}
        \caption{Subjective Evaluation Criteria for Activity-Travel Chains} \\
		\label{tab:evaluation_criteria} \\
        \toprule
        \textbf{Dimension} & \textbf{Score Range} & \textbf{Criteria and Requirements} \\
        \midrule
        \endfirsthead
        
        \toprule
        \textbf{Dimension} & \textbf{Score Range} & \textbf{Criteria and Requirements} \\
        \midrule
        \endhead
        
        \textbf{Time Logic Consistency (Weight: 25\%)} & \centering 9-10 points & \textbf{strong time logic, optimal utility allocation.} Activity sequence completely reasonable with no time conflicts; Time allocation highly realistic, conforming to life patterns; Sufficient and feasible transition time between activities; Fully reflects individual utility maximization under time constraints. \\
        
        \textbf{} & \centering 7-8 points & \textbf{Good time logic, reasonable utility allocation.} Activity sequence basically reasonable with occasional minor time pressure; Time allocation relatively realistic; Transition time between activities basically feasible; Better reflects time utility optimization principles. \\
        
        \textbf{} & \centering 5-6 points & \textbf{Average time logic, problematic utility allocation.} Activity sequence has certain unreasonableness; Partial time allocation insufficiently realistic; Transition time between activities too tight or too loose; Time utility allocation needs optimization. \\
        
        \textbf{} & \centering 3-4 points & \textbf{Poor time logic, inappropriate utility allocation.} Activity sequence unreasonable in multiple places; Time allocation obviously deviates from reality; Infeasible transition time between activities; Fails to reflect reasonable time utility consideration. \\
        
        \textbf{} & \centering 1-2 points & \textbf{Chaotic time logic, no utility optimization.} Severely disordered activity sequence; Completely unrealistic time allocation; Obvious time conflicts exist; Completely ignores time utility maximization principles. \\
        
        \midrule
        
        \textbf{Activity Purpose Coherence (Weight: 25\%)} & \centering 9-10 points & \textbf{Excellent purpose coherence, optimal utility trade-off.} Strong logical correlation in activity sequence; Perfect match between travel purpose and activity types; Highly reasonable overall activity chain; Fully reflects utility trade-off optimization in multi-objective decisions. \\
        
        \textbf{} & \centering 7-8 points & \textbf{Good purpose coherence, reasonable utility trade-off.} Strong logical correlation in activity sequence; High match between travel purpose and activity types; Overall activity chain relatively reasonable; Better reflects utility trade-off mechanisms. \\
        
        \textbf{} & \centering 5-6 points & \textbf{Average purpose coherence, flawed utility trade-off.} Average logical correlation in activity sequence; Medium match between travel purpose and activity types; Activity chain has certain unreasonableness; Insufficient reflection of utility trade-off mechanisms. \\
        
        \textbf{} & \centering 3-4 points & \textbf{Poor purpose coherence, imbalanced utility trade-off.} Weak logical correlation in activity sequence; Low match between travel purpose and activity types; Overall activity chain insufficiently reasonable; Lacks effective utility trade-off consideration. \\
        
        \textbf{} & \centering 1-2 points & \textbf{Very poor purpose coherence, no utility trade-off.} Activity sequence lacks logical correlation; Serious mismatch between travel purpose and activity types; Overall chaotic activity chain; Completely ignores utility trade-off principles. \\
        
        \midrule
        
        \textbf{Persona Matching Degree (Weight: 25\%)} & \centering 9-10 points & \textbf{Extremely high persona matching, accurate utility preferences.} Activity patterns highly match persona characteristics; Sufficient reflection of occupation, age, lifestyle habits characteristics; Perfect reflection of typical utility preferences of the group; Outstanding and reasonable personalized characteristics. \\
        
        \textbf{} & \centering 7-8 points & \textbf{Good persona matching, relatively accurate utility preferences.} Activity patterns match persona characteristics well; Main characteristics clearly reflected; Better reflection of group utility preference characteristics; Basically reasonable personalized characteristics. \\
        
        \textbf{} & \centering 5-6 points & \textbf{Average persona matching, vague utility preferences.} Medium matching between activity patterns and persona characteristics; Insufficient reflection of some characteristics; General reflection of group utility preferences; Insufficiently prominent personalized characteristics. \\
        
        \textbf{} & \centering 3-4 points & \textbf{Poor persona matching, incorrect utility preferences.} Low matching between activity patterns and persona characteristics; Most characteristics not obviously reflected; Inappropriate reflection of group utility preferences; Missing or unreasonable personalized characteristics. \\
        
        \textbf{} & \centering 1-2 points & \textbf{Very poor persona matching, no utility preference reflection.} Activity patterns seriously inconsistent with persona characteristics; Chaotic or missing characteristic reflection; Completely ignores group utility preferences; No reasonable personalization. \\
        
        \midrule
        
        \textbf{Activity Authenticity (Weight: 25\%)} & \centering 9-10 points & \textbf{Extremely high activity authenticity, optimal rational choice.} Activity content completely conforms to reality; Highly reasonable duration; Optimal location selection; Fully reflects rational choice behavior under resource constraints. \\
        
        \textbf{} & \centering 7-8 points & \textbf{Good activity authenticity, reasonable rational choice.} Activity content basically conforms to reality; Relatively reasonable duration; Appropriate location selection; Better reflects rational choice principles. \\
        
        \textbf{} & \centering 5-6 points & \textbf{Average activity authenticity, biased rational choice.} Activity content roughly conforms to reality; Duration has certain deviations; Average location selection; Insufficient reflection of rational choice. \\
        
        \textbf{} & \centering 3-4 points & \textbf{Poor activity authenticity, inappropriate rational choice.} Activity content deviates from reality; Obviously unreasonable duration; Inappropriate location selection; Lacks rational choice consideration. \\
        
        \textbf{} & \centering 1-2 points & \textbf{Very poor activity authenticity, no rational choice.} Activity content seriously deviates from reality; Completely unreasonable duration; Absurd location selection; Completely ignores rational choice behavior. \\
        
        \bottomrule
    \end{longtable}
}

\clearpage
\subsection{Reasoning Capabilities Evaluation Criteria}
\label{Reasoning Capabilities Evaluation Criteria}

{
\renewcommand{\arraystretch}{1.05} 
\begin{table}[htbp]
    \centering
    \caption{Reasoning Capabilities Evaluation Dimensions and Scoring Criteria}
    \label{tab:reasoning_evaluation_criteria}
    \begin{tabular}{p{2.2cm} c p{8cm} c}
        \toprule
        \textbf{Evaluation Dimension} & \textbf{Weight} & \textbf{Scoring Criteria} & \textbf{Score Range} \\
        \midrule
        Information Collection Completeness & 20\% & 
        \textbf{Excellent (8-10):} Proactively identifies and collects all critical information dimensions including temporal constraints, spatial positioning, user preferences, resource limitations, and environmental factors\newline
        \textbf{Good (6-7):} Collects basic information with minor omissions in important dimensions\newline
        \textbf{Fair (4-5):} Insufficient information collection with multiple critical gaps\newline
        \textbf{Poor (0-3):} Minimal or ineffective information gathering & 0-10 \\
        \midrule
        
        Analytical Thinking Depth & 20\% & 
        \textbf{Excellent (8-10):} Conducts comprehensive multi-dimensional analysis covering temporal, spatial, cost-benefit, risk assessment, and personalization aspects with consideration of complex interrelationships\newline
        \textbf{Good (6-7):} Performs basic analysis with insufficient depth\newline
        \textbf{Fair (4-5):} Superficial analysis lacking thorough consideration\newline
        \textbf{Poor (0-3):} Minimal or ineffective analytical processes & 0-10 \\
        \midrule
        
        Logical Reasoning Rigor & 20\% & 
        \textbf{Excellent (8-10):} Demonstrates rigorous logical reasoning with clear causal relationships, valid premises, and well-supported conclusions throughout each reasoning step\newline
        \textbf{Good (6-7):} Generally sound logic with some apparent reasoning deficiencies\newline
        \textbf{Fair (4-5):} Contains logical gaps or contradictions\newline
        \textbf{Poor (0-3):} Confused logic with unreasonable reasoning processes & 0-10 \\
        \midrule
        
        Decision-Making Rationality & 20\% & 
        \textbf{Excellent (8-10):} Clearly identifies conflicts, demonstrates transparent and rational trade-off processes, provides well-justified decisions with adequate consideration of alternatives\newline
        \textbf{Good (6-7):} Insufficient consideration of trade-offs with unclear decision justification\newline
        \textbf{Fair (4-5):} Limited conflict identification with superficial balancing processes\newline
        \textbf{Poor (0-3):} Absence of trade-off processes with unjustified decisions & 0-10 \\
        \midrule
        
        Tool Invocation Strategy & 20\% & 
        \textbf{Excellent (8-10):} Perfect timing of tool calls with reasonable sequencing and strong strategic planning\newline
        \textbf{Good (6-7):} Generally appropriate tool usage with occasional improper applications\newline
        \textbf{Fair (4-5):} Obvious problems in tool utilization lacking strategic approach\newline
        \textbf{Poor (0-3):} Chaotic or missing tool usage & 0-10 \\
        \bottomrule
    \end{tabular}
\end{table}
\vspace{2pt}
\small \textbf{Note:} Final scores are calculated as weighted averages across all five dimensions, each contributing 20\% to the overall reasoning quality assessment.
}

\end{document}